\documentclass[letterpaper]{article} 
\usepackage{aaai2027}
\nocopyright
\usepackage[hyphens]{url}  
\usepackage{graphicx} 
\usepackage{natbib}  
\usepackage{caption} 
\usepackage{amsmath}
\usepackage{amssymb}
\usepackage{algorithm}
\usepackage{algorithmic}
\usepackage{booktabs}
\usepackage{multirow}
\usepackage{graphicx}
\usepackage{newfloat}
\usepackage{listings}
\DeclareCaptionStyle{ruled}{labelfont=normalfont,labelsep=colon,strut=off} 
\floatstyle{ruled}
\newfloat{listing}{tb}{lst}{}
\floatname{listing}{Listing}

\usepackage{booktabs}

\title{Think with Extra-Image: A Farmland Segmentation Agent Driven by Spatio-Temporal Information Gain}

\author {
    Haiyang Wu\textsuperscript{\rm 1}\equalcontrib,
    Weiliang Mu\textsuperscript{\rm 1}\equalcontrib,
    Zhuofei Du\textsuperscript{\rm 1},
    Dandan Zhong\textsuperscript{\rm 1},
    Kaijie Shi\textsuperscript{\rm 1},
    Haifeng Li\textsuperscript{\rm 1},
    Chao Tao\textsuperscript{\rm 1}\corresponding
}
\affiliations {
    \textsuperscript{\rm 1}Central South University, Changsha, China\\
    ocean-w@csu.edu.cn, 245012170@csu.edu.cn, kingtaochao@csu.edu.cn
}

\begin{document}

\maketitle

\begin{abstract}
Existing farmland remote sensing image (FRSI) segmentation follows a “Think with Intra-Image” paradigm, assuming that the current image contains sufficient visual evidence for reliable segmentation. Yet farmland appearance varies with phenology and spatial context and is often confused with other land-cover, making instantaneous, local observations inadequate. Thus, segmentation ambiguity stems not only from limited model representation, but more fundamentally from the required spatio-temporal information lying beyond the current image. Based on this insight, we redefine FRSI segmentation from an information bottleneck perspective as a dynamic decision process driven by task-relevant extra spatio-temporal information gain. We further propose FarmSeeker, a dynamic FRSI segmentation agent that identifies ambiguous regions, reasons about their causes, and queries extra spatio-temporal information on demand for accurate segmentation. To evaluate FarmSeeker, we construct GSFS-Bench, the first global-scale, high-resolution FRSI segmentation benchmark that supports reasoning-querying. Experiments show that FarmSeeker achieves more stable segmentation performance than existing methods. The project is publicly available at: https://withoutocean.github.io/FarmSeeker/

\end{abstract}


\section{Introduction}

Farmland remote sensing image (FRSI)  segmentation has long been regarded as a representation learning problem: given an RSI, the model is expected to follow a “Think with Intra-Image” paradigm, extracting discriminative features from the current image to recover a pixel-level farmland mask \cite{zheng2026comprehensive, wang2026comprehensive, hossain2026vision}. However, this paradigm implicitly assumes that the current image already contains sufficient visual evidence for FRSI segmentation. This raises a conjecture: could the key bottleneck in FRSI segmentation lie not in insufficient model representation capacity, but in the incomplete visual evidence provided by the current image? 

The key basis for this view lies in the inherent spatio-temporal dependence of farmland \cite{zhang2025gtpbd}. Temporally, the farmland status of a parcel may remain unchanged while its visual appearance varies across phenological stages, exhibiting states such as vegetation cover, bare soil, or surface water. In other words, the same farmland may present substantially different visual appearances over time \cite{zhu2025exact}. Spatially, farmland recognition depends on boundary continuity and contextual relationships with neighboring land-cover types, such that similar appearances may correspond to different classes under different spatial configurations \cite{muhawenayo2026prue}. Therefore, many errors in complex scenes do not fundamentally arise because the model fails to extract sufficient information from the current image, but because the critical evidence required for reliable recognition may lie beyond the observational scope of that image.

When the current image does not contain the complete chain of spatio-temporal evidence required for farmland recognition, a natural solution is to actively acquire the missing evidence from external observations on demand. The human interpretation paradigm provides important insights. When experts encounter an ambiguous region that appears inundated in the current image, they do not draw a conclusion solely from the current observation. Instead, they retrieve images of the same parcel from other periods to determine whether it is a paddy field during the irrigation stage. When a locally cropped image happens to truncate field boundaries, experts expand the observation range and incorporate spatial relationships into their assessment. In this process, humans do not simply inspect more images. Instead, guided by uncertainty in the current observation, they examine external spatio-temporal data on demand to fill critical information gaps.

Based on the above view, we first theoretically clarify the necessity of actively acquiring external spatio-temporal information gain, and accordingly propose FarmSeeker, a farmland segmentation agent framework driven by spatio-temporal information gain. The framework is organized around four core questions: when external information is needed, what information is required, how it should be acquired, and how the newly obtained evidence should be utilized. To implement this framework, we construct FM-Seg69K, a multi-stage training dataset, and design a corresponding progressive training strategy (PTS) and task-driven reward functions (TRFs). Specifically, FM-Seg69K is constructed around a complete task chain comprising ambiguity perception, evidence need assessment, tool calling, cross spatio-temporal collaborative reasoning, and mask refinement. The PTS organizes this task chain into a staged learning process, enabling the model to acquire the required capabilities in a stable and progressive manner. The TRFs translate the key decisions and final segmentation results into explicit optimization signals, guiding the model toward effective information acquisition and utilization. To evaluate FarmSeeker, we further establish GSFS-Bench, the global-scale, high-resolution FRSI segmentation benchmark that supports dynamic querying and cross spatio-temporal reasoning. Experiments show that FarmSeeker achieves more stable segmentation performance across diverse regions and ambiguous scenarios.

\begin{itemize}
    \item Motivated by an information-bottleneck view, we formulate farmland segmentation under incomplete observations as an on-demand extra-image evidence acquisition problem and introduce the Think with Extra-Image paradigm.
    \item We propose FarmSeeker, which constructs a multi-stage training dataset FM-Seg69K and designs a PTS along with TRFs, enabling FarmSeeker to develop a closed-loop capability for on-demand acquisition of spatio-temporal information gain.
    \item We construct GSFS-Bench, a global high-resolution farmland segmentation benchmark that supports the evaluation of in-region and cross-region generalization, ambiguity perception, tool calling, and multi-image collaborative reasoning.
\end{itemize}

\section{Related Work}

\textbf{Applications of Extra Spatio-Temporal Information in FRSI Segmentation.} Existing studies have widely integrated multi-temporal information and spatial context to improve segmentation performance \cite{liu2025global,wang2025joint,xie2026cnn,li2026agrifm}. Temporal fusion methods model farmland dynamics across seasons, mitigating semantic ambiguity caused by transient appearances \cite{lobert2025unveiling}. Spatial-context methods leverage field morphology, boundary continuity, and neighborhood relationships to improve discrimination in complex scenes \cite{li2024comprehensive}. However, these methods typically treat extra spatio-temporal data as predefined inputs for passive fusion, potentially weakening the discriminative contribution of key evidence \cite{turkoglu2026t3sthinkthermaltime,zhao2026weakly}.

\textbf{Active Visual Reasoning and Remote-Sensing Agents.} Recent studies on agents have shown that models can supplement visual evidence through active re-observation and tool calling, and update their judgments accordingly \cite{li2025designingdomainspecificagentshierarchical,yao2026remoteagent,yao2026remotereasoner}. For example, DeepEyes \cite{zheng2026deepeyes} can fill evidence gaps in the current observation through active visual revisitation. ZoomEarth \cite{liu2026zoomearth} first establishes a global understanding of the scene, then adaptively localizes key regions and performs fine-grained re-observation through cropping and zooming. Although these works demonstrate the feasibility of information-seeking visual reasoning, their evidence acquisition remains largely confined to the current image, following a “Think with Intra-Image” \cite{shao2026asking}. Moreover, existing methods primarily target image understanding and visual question answering, and lack a complete mechanism for translating actively acquired evidence into pixel-level segmentation refinement.

\section{Why Extra-Image Spatio-Temporal Information Gain is Needed}
\begin{figure}[t]
\centering
\includegraphics[width=0.9\columnwidth]{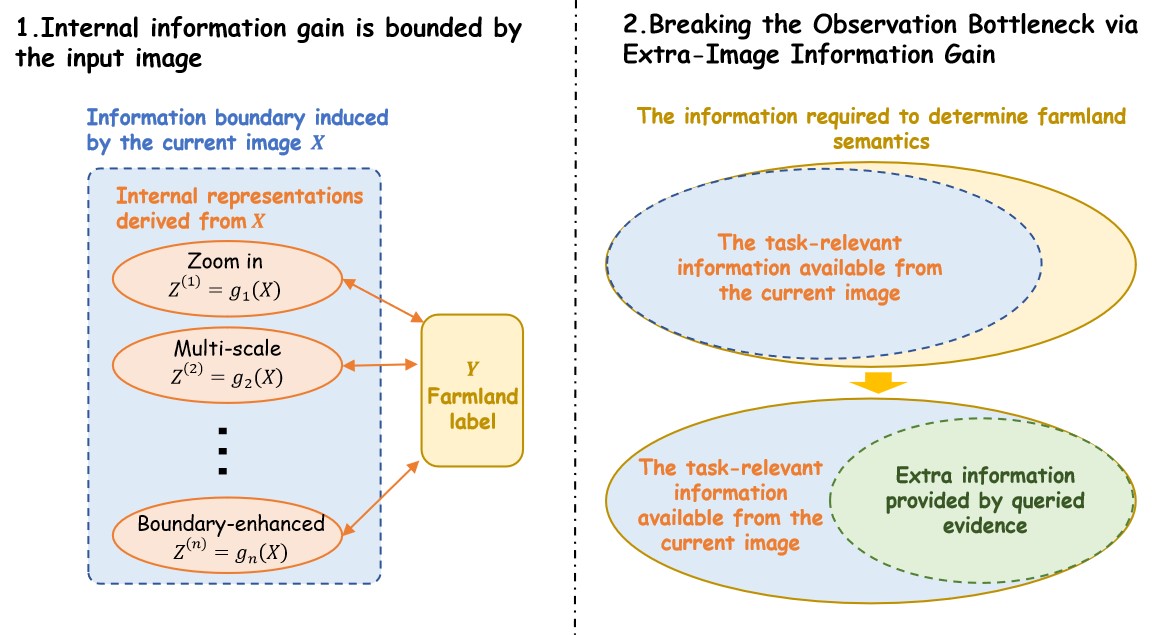} 
\caption{Conceptual illustration of intra-image and extra-image spatio-temporal information gain.}
\label{fig2}
\end{figure}

Given a farmland remote sensing image (FRSI) $X$, the ground-truth farmland semantics $Y$, and an internal representation $Z$ extracted from $X$, the information bottleneck \cite{tishby2000informationbottleneckmethod,hu2024survey} aims to learn a compact yet task-relevant representation by optimizing $\max_{p(z\mid x)}[I(Z;Y)-\beta I(Z;X)]$, where $I(\cdot;\cdot)$ denotes mutual information, $I(Z;Y)$ measures the farmland-discriminative information retained in $Z$, $I(Z;X)$ measures the input information preserved by the representation, and $\beta>0$ controls the trade-off between task relevance and representation compression. Since $Z$ is derived solely from $X$, the variables form the Markov chain $Y\rightarrow X\rightarrow Z$. According to the data processing inequality, $I(Z;Y)\leq I(X;Y)$. This implies that multi-scale modeling, zoom-in operations, and boundary enhancement can only better extract and organize the information already contained in the current image, but cannot generate discriminative evidence that is absent from it. As illustrated in Figure~\ref{fig2}, these operations can be expressed as internal transformations $Z^{(k)}=g_k(X)$. The right of Figure~\ref{fig2} further illustrates this limitation from a task-information perspective. The yellow region denotes the information required to determine farmland semantics, $H(Y)$, whereas the blue region denotes the task-relevant information available from the current image, $I(Y;X)$. Phenological changes may cause farmland to exhibit visual characteristics similar to bare soil, vegetated land, or water bodies during specific periods, while local image windows may truncate neighborhood structural information. In such cases, $I(Y;X)$ is insufficient to support a reliable prediction of $Y$. The primary limitation is therefore no longer only the model's representation capacity, but a more fundamental observation bottleneck.

To this end, let $q$ denote a query action
and $E_q$ the extra-image evidence returned by the query, such
as imagery from another time point or a broader spatial context.
After incorporating $E_q$, the available task-relevant information
satisfies
\begin{equation}
I(Y;X,E_q)
=
I(Y;X)+I(Y;E_q\mid X)
\label{eq:information_decomposition}
\end{equation}
Here, the conditional mutual information $I(Y;E_q\mid X)$ quantifies the extra task-relevant information provided by $E_q$ given the current image $X$. It corresponds to the green region in the right of Figure~\ref{fig2}. Unlike internal representations derived from $X$, $E_q$ originates beyond the current observation and can therefore extend the observable information boundary. Accordingly, we define the task-relevant information gain of query
$q$ as
\begin{equation}
I(Y;E_q\mid X)
=
H(Y\mid X)-H(Y\mid X,E_q)
\label{eq:information_gain}
\end{equation}
where $H(Y\mid X)$ denotes the remaining uncertainty in farmland semantics given only the current image, and $H(Y\mid X,E_q)$ denotes the uncertainty after incorporating the queried evidence.

Temporal and spatial queries address different information deficiencies. Extra temporal imagery provides phenological evolution cues for resolving temporal ambiguity, whereas broader spatial imagery provides boundary continuity and neighborhood organization for resolving spatial ambiguity. The objective is therefore not to indiscriminately introduce more images, but to select the query that provides the largest task-relevant information gain:
\begin{equation}
q^{*}
=
\mathop{\mathrm{arg\,max}}_{q\in\mathcal{Q}}
I(Y;E_q\mid X)
\label{eq:optimal_query}
\end{equation}
where $\mathcal{Q}$ denotes the set of candidate spatio-temporal
queries and $q^{*}$ denotes the optimal query action. Equation~(3) defines an ideal query objective rather than a directly computable training target. In FarmSeeker, this objective is not explicitly optimized. Instead, it is implicitly approximated through  a learned query policy that identifies insufficient observations, predicts the type of missing evidence, and selects an appropriate spatio-temporal query under task-oriented supervision and outcome-based rewards.

\section{GSFS-Bench}
\begin{figure}[t]
\centering
\includegraphics[width=0.9\columnwidth]{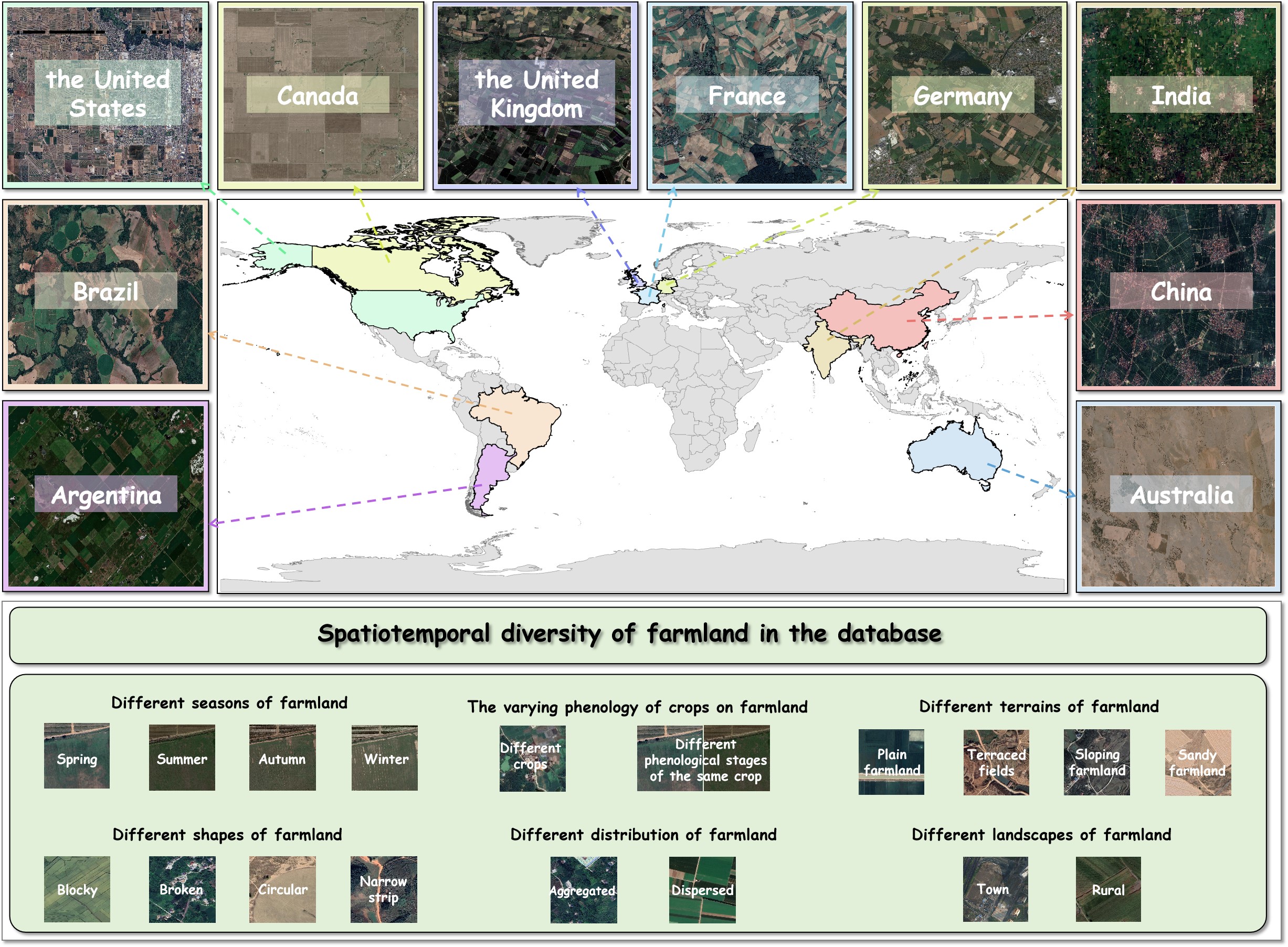} 
\caption{spatio-temporal diversity of the GSFS-Bench.}
\label{fig3}
\end{figure}

Existing high-resolution farmland segmentation benchmarks cannot adequately capture substantial distribution shifts across global agricultural landscapes. They also cannot assess whether a model can identify missing information, actively acquire additional imagery, and use the newly obtained evidence to revise its segmentation results. 

As illustrated in Fig.~\ref{fig3}, we construct GSFS-Bench, which covers representative regions from more than ten major agricultural-producing countries. For each region, we collect multi-temporal FRSIs from Google Earth within a time span of no more than one year. In total, the benchmark contains more than 200 wide-swath images with spatial resolutions ranging from 0.5 to 1~m. After manually removing areas affected by cloud cover, experts provide pixel-level farmland annotations. Subsequently, one temporal image from each region is selected as the current observation and uniformly cropped into $512 \times 512$ pixels, resulting in a segmentation evaluation set containing 10K samples. The remaining temporal images and their corresponding original wide-swath images jointly constitute queryable temporal and spatial evidence pools. To support different levels evaluation, we divide the regions in China into eight agricultural regions according to their agro-geographical characteristics for in-region evaluation (The training samples used in this work are strictly separated from GSFS-Bench). Regions outside China are used for cross-region evaluation. Furthermore, to evaluate ambiguity perception, tool calling, and collaborative reasoning, we construct three evaluation subsets based on the data from China: ambiguity perception, tool-calling type classification, and multi-image collaborative reasoning. \textbf{More details are provided in the supplementary material.}

\section{FarmSeeker}

\subsection{Overview}
\begin{figure*}[t]
\centering
\includegraphics[width=0.8\textwidth]{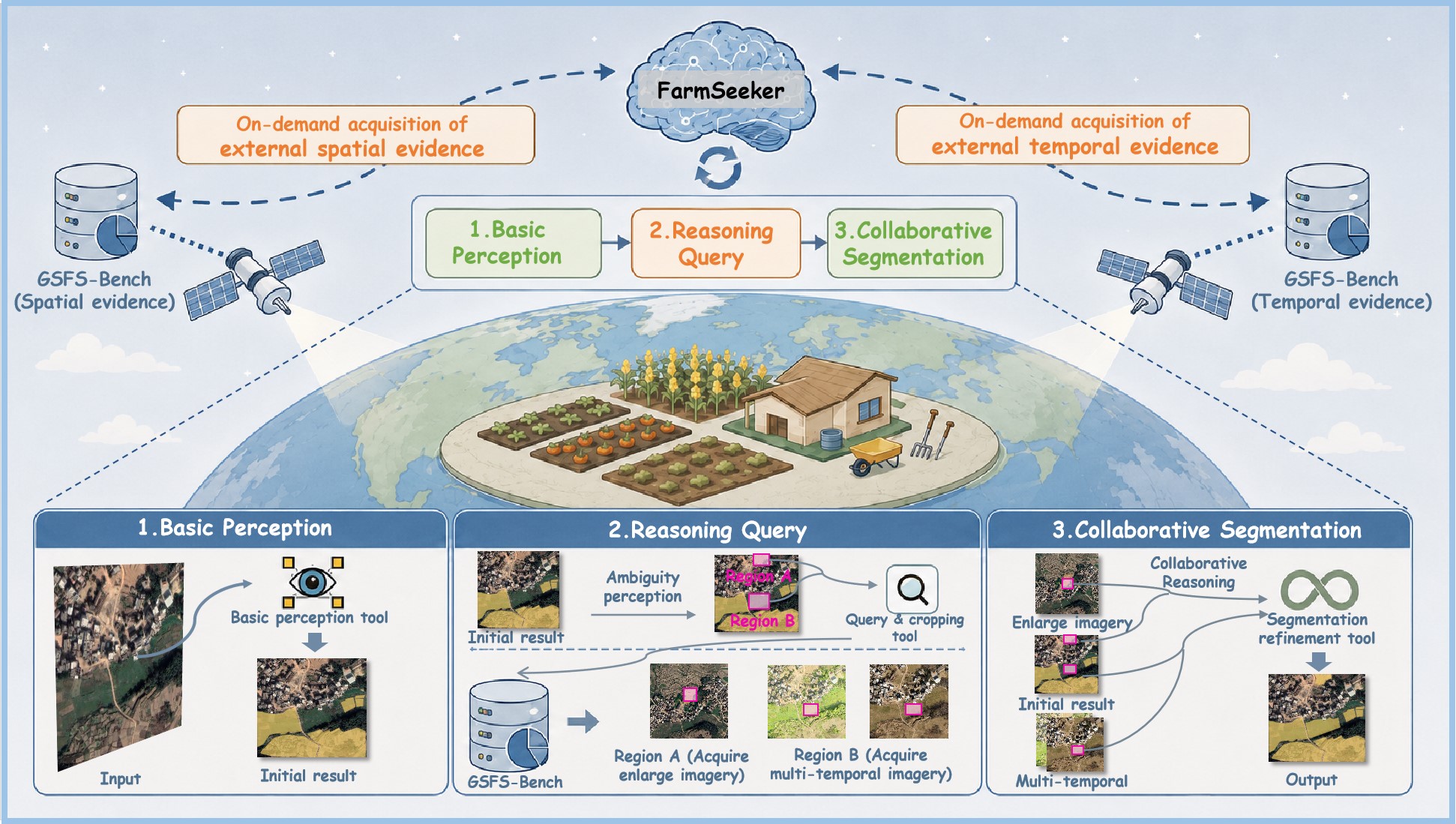} 
\caption{Overview of FarmSeeker. The reasoning engine and tools collaborate to complete three stages: \textbf{Basic Perception} calls a basic perception tool to generate an initial farmland mask; \textbf{Reasoning Query} performs ambiguity perception and evidence need assessment, and accordingly calls query and cropping tools to acquire extra evidence; \textbf{Collaborative Segmentation} performs collaborative reasoning over the initial prediction and retrieved evidence, and calls a segmentation refinement tool to produce the final mask.}
\label{fig4}
\end{figure*}

As shown in Fig.~\ref{fig4}, FarmSeeker is a dynamic FRSI segmentation framework composed of a multimodal large language model (MLLM) based reasoning engine (RE) $\mathcal{R}$ and a tool repository containing a basic perception tool, a query-and-cropping tool, and a segmentation refinement tool. It consists of three stages: basic perception, reasoning query, and collaborative segmentation. Given an FRSI $X$, the basic perception tool $\mathcal{T}_p$ first generates an initial farmland mask:
\begin{equation}
M_0=\mathcal{T}_p(X).
\label{eq:initial_mask}
\end{equation}
The RE then jointly analyzes $X$ and $M_0$ to localize
ambiguous regions $b_i$ with insufficient evidence and determine a query strategy $q_i\in\{\mathrm{temporal},\mathrm{spatial},\mathrm{none}\}$
for each region. For regions requiring extra information, the query tool $\mathcal{T}_q$ retrieves extra-image evidence $E_i$ from the spatio-temporal evidence pool. The RE further integrates the input image, the initial mask and the newly acquired evidence to generate a regional-level refinement decision $s_i$. Finally, the segmentation refinement tool $\mathcal{T}_s$ converts each decision at the region-level into a prediction at the local pixel-level and fuses it with the initial mask:
\begin{equation}
M^{*}
=
\mathop{\mathrm{Fuse}}
\left(
M_0,
\{\mathcal{T}_s(X,b_i,s_i)\}_{i=1}^{N}
\right),
\label{eq:final_mask}
\end{equation}
where $M^{*}$ denotes the final result of the FRSI segmentation. 


\subsection{FM-Seg69K}
To support the training of FarmSeeker, we construct FM-Seg69K, which
consists of four parts. \textbf{The basic perception data} uses FarmSeg-VL \cite{tao2025large} to train the basic perception tool. \textbf{The General Fine-Tuning (FT) data} are constructed by filtering and uniformly organizing existing datasets, including EarthReason \cite{li2025segearthr1geospatialpixelreasoning}, FarmSeg-VL, WeThink \cite{yang2025wethinkgeneralpurposevisionlanguagereasoning}, and DIOR-RSVG \cite{zhan2023rsvg}, to enhance general understanding, region localization, and instruction following capabilities. \textbf{The Cold-Start FT data} and \textbf{Reinforcement Fine-Tuning (RFT) data} are independently constructed around the core capabilities of FarmSeeker. The former uses initial segmentation results and ground-truth annotations to construct samples for ambiguous-region localization, evidence need assessment, tool calling, and cross spatio-temporal collaborative reasoning, thereby initializing the task behaviors of the reasoning engine. The latter adopts the same task format but further improves sample quality and difficulty hierarchy. \textbf{More details are provided in the supplementary material.}

\subsection{Progressive Training Strategy}

Inspired by recent post-training paradigms \cite{zhao2026chemdfmrchemicalreasoningllm}, we propose a progressive training strategy. We sequentially apply General FT, Cold-Start FT, and RFT to train the RE. General FT first establishes remote-sensing scene understanding, region localization, and general reasoning capabilities, providing a foundation for identifying anomalous regions in images and masks. Cold-Start FT then introduces ambiguity localization, spatio-temporal query planning, and multi-image collaborative reasoning through structured task trajectories, providing a stable initialization for RFT. Finally, RFT combines GRPO \cite{guo2025deepseek} with task-driven reward functions to further optimize ambiguity perception, tool calling, and evidence utilization.

\subsection{Task-driven reward functions}
FarmSeeker involves two capabilities: ambiguity perception and tool calling $T_{\mathrm{det}}$, and collaborative extra-image reasoning $T_{\mathrm{rea}}$. Since the two capabilities have different objectives, we design task-specific rule-based rewards.

\textbf{Ambiguity perception and Tool calling.} This capability requires the RE to localize ambiguous regions, produce valid bounding boxes, and select an appropriate query tool for each detected region. We evaluate these objectives through format validity, detection coverage, detection efficiency, and tool calling accuracy. A detection output is valid when each \texttt{bbox} is paired with a \texttt{label} and contains four coordinates. Let $f_{\mathrm{det}}=1$ denote a valid format. The detection format reward is defined as
\begin{equation}
R_{\mathrm{fmt}}^{\mathrm{det}}y
=
\left\{
\begin{array}{rl}
0,  & f_{\mathrm{det}}=1,\\
-1, & \mathrm{otherwise}
\end{array}
\right.
\label{eq:det_format_reward}
\end{equation}
This negative-only constraint prevents the model from over-optimizing simple formatting rules. We then apply Hungarian matching \cite{carion2020end} between the predicted and ground-truth boxes. Let $TP$, $GT$, and $Pred$ denote the numbers of valid detections, ground-truth boxes, and predicted boxes, respectively. The detection recall reward is
\begin{equation}
R_{\mathrm{rec}}
=
\left\{
\begin{array}{ll}
\displaystyle \frac{TP}{GT}, & GT \geq 1, \\[6pt]
0,                            & GT = 0
\end{array}
\right.
\label{eq:recall_reward}
\end{equation}
To discourage excessive predictions, the detection efficiency reward is defined as 
\begin{equation}
R_{\mathrm{eff}}
=
\left\{
\begin{array}{ll}
\displaystyle \frac{TP}{Pred}, & Pred \geq 1, \\[6pt]
0, & Pred = 0
\end{array}
\right.
\label{eq:efficiency_reward}
\end{equation}
Let $\mathcal{V}$ denote the set of valid matches, and let $\hat{u}_i$ and $u_i$ denote the predicted and ground-truth tool types for the $i$-th matched region. The tool calling reward is
\begin{equation}
R_{\mathrm{tool}}
=
\left\{
\begin{array}{ll}
\displaystyle
\frac{1}{|\mathcal{V}|}
\sum_{i\in\mathcal{V}}
\mathbf{1}\left(\hat{u}_i=u_i\right),
& |\mathcal{V}| \geq 1, \\[8pt]
0, & |\mathcal{V}| = 0
\end{array}
\right.
\label{eq:tool_reward}
\end{equation}
These rewards encourage complete ambiguity localization, suppress redundant bounding boxes, and establish a reliable mapping between ambiguity types and query tools.

\textbf{Collaborative Reasoning.} This capability requires the RE to integrate the queried evidence and predict the category of the target region. Given the predicted label $\hat{y}$ and the ground-truth label $y$, the answer reward is
\begin{equation}
R_{\mathrm{ans}}
=
\mathbf{1}(\hat{y}=y)
\label{eq:answer_reward}
\end{equation}
To avoid rewarding correct answers obtained through accidental guessing, we employ Qwen3-32B as a fixed LLM judge. Given the question, reasoning trajectory, and final answer, it evaluates answer consistency $r_{\mathrm{con}}$, linguistic and logical coherence $r_{\mathrm{gram}}$, and task compliance $r_{\mathrm{traj}}$. The reasoning reward is defined as
\begin{equation}
R_{\mathrm{cot}}
=
\left\{
\begin{array}{rl}
1, &
r_{\mathrm{con}}=r_{\mathrm{gram}}=r_{\mathrm{traj}}=1,\\
0, & \mathrm{otherwise}.
\end{array}
\right.
\label{eq:cot_reward}
\end{equation}
The complete judge prompt is provided in the supplementary material. The model is also required to place its reasoning and final prediction within \texttt{<think>\allowbreak</think>} and \texttt{<answer>\allowbreak</answer>}, respectively. Let $f_{\mathrm{think}}=1$ denote a valid output structure. The reasoning format reward is
\begin{equation}
R_{\mathrm{fmt}}^{\mathrm{think}}
=
\left\{
\begin{array}{rl}
0,  & f_{\mathrm{think}}=1,\\
-1, & \mathrm{otherwise}
\end{array}
\right.
\label{eq:reasoning_format_reward}
\end{equation}

\textbf{Task-Driven Reward Aggregation.} For ambiguity perception and tool calling, the total reward is
\begin{equation}
R_{\mathrm{det}}
=
R_{\mathrm{fmt}}^{\mathrm{think}}
+
R_{\mathrm{fmt}}^{\mathrm{det}}
+
R_{\mathrm{rec}}
+
R_{\mathrm{eff}}
+
R_{\mathrm{tool}}
\label{eq:detection_total_reward}
\end{equation}
For collaborative extra-image reasoning, the total reward is
\begin{equation}
R_{\mathrm{rea}}
=
R_{\mathrm{fmt}}^{\mathrm{think}}
+
R_{\mathrm{cot}}
+
R_{\mathrm{ans}}
\label{eq:reasoning_total_reward}
\end{equation}
Accordingly, the task-driven reward for response $o_i$ generated from
sample $x_i$ is defined as
\begin{equation}
R(o_i)
=
\left\{
\begin{array}{ll}
R_{\mathrm{det}}, & x_i\in T_{\mathrm{det}},\\
R_{\mathrm{rea}}, & x_i\in T_{\mathrm{rea}}
\end{array}
\right.
\label{eq:unified_reward}
\end{equation}
Through the above design, the RE can jointly optimize ambiguity perception, tool calling, and tool-feedback-based collaborative reasoning during RFT.

\section{Experiments}
 The basic perception tool of FarmSeeker and all compared methods are trained on the same FarmSeg-VL subset of FM-Seg69K. The compared methods include DeepLabv3+ \cite{chen2018encoder}, DDRNet \cite{pan2022deep}, DSNet \cite{guo2024dsnet}, DBBANet \cite{li2024comprehensive}, LaSagnA(7B) \cite{wei2024lasagnalanguagebasedsegmentationassistant}, PixelLM(7B) \cite{ren2024pixellm}, SegEarth-R1(1.3B) \cite{li2025segearthr1geospatialpixelreasoning}, and FSVLM(7B) \cite{wu2025fsvlm}. FarmSeeker uses SegEarth-R1 as the basic perception tool, a fine-tuned Qwen2.5-VL-7B \cite{bai2025qwen25vltechnicalreport} as the reasoning engine, and \texttt{sam2.1\_hiera\_large.pt} \cite{ravi2024sam2segmentimages} as the segmentation refinement tool. Unless otherwise specified, all experiments in the main text are conducted on the mixed test subset comprising eight major agricultural regions in China from GSFS-Bench. The main text reports only Recall and IoU. All experiments are conducted on four A800. \textbf{More experimental results are provided in the supplementary material.} 

\subsection{In-Region and Cross-Region Performance}

Table~\ref{tab1} reports the IoU of all methods across 8 agricultural regions in China (in-region) and 11 agricultural countries outside China (cross-region). FarmSeeker achieves the best results in 6 in-region agricultural regions and ranks second in the remaining 2. It also achieves the best results in 10 cross-region countries, demonstrating stable in-region performance and strong cross-region generalization. Further analysis shows that the gains of FarmSeeker are closely related to regional difficulty. In regions where baseline performance is already high, such as the Northeast China Plain, Huang-Huai-Hai Plain, Germany, and Canada, the improvements are relatively limited. In contrast, more substantial gains are observed in more challenging regions, including South China Areas, the Sichuan Basin, Australia, and Argentina, indicating that extra spatio-temporal evidence is particularly beneficial when the current image provides insufficient information.

\begin{table*}[t]
\centering
\resizebox{\textwidth}{!}{
\begin{tabular}{lccccccccc}
\toprule
Region / Country
& DeepLabv3+ & DDRNet & DSNet & DBBANet
& LaSagnA & PixelLM & FSVLM & SegEarth-R1 & FarmSeeker \\

\midrule
\multicolumn{10}{c}{\textit{In-Region}} \\
\midrule

Northeast China Plain
& 94.81 & 92.57 & 94.19 & 95.06
& 94.16 & 94.57 & 93.93 & \underline{95.47}
& \textbf{95.99} \\

Huang-Huai-Hai Plain
& 95.73 & 91.38 & 95.28 & \underline{96.50}
& 94.27 & 95.56 & 94.31 & 96.38
& \textbf{96.78} \\

Northern Arid and Semi-arid Region
& 85.90 & 85.91 & 82.63 & 88.24
& \underline{88.58} & 87.68 & 87.53 & 85.64
& \textbf{88.73} \\

Loess Plateau
& 94.72 & 89.40 & 91.93 & 90.77 
& 85.16 & 95.13 & \textbf{95.96} & 94.78
& \underline{95.49} \\

South China Areas
& 65.14 & 63.87 & 59.17 & 71.15
& 53.57 & 69.36 & 69.39 & \underline{73.01}
& \textbf{75.23} \\

Sichuan Basin
& 77.23 & 76.64 & 77.15 & 79.17
& 79.25 & \underline{79.86} & 78.57 & 78.30
& \textbf{81.57} \\

Yunnan-Guizhou Plateau
& 81.20 & 66.21 & 81.46 & 80.74
& 73.96 & 78.76 & \underline{85.03} & 84.73
& \textbf{86.41} \\

Middle-lower Yangtze Plain
& 75.13 & 75.20 & 74.52 & 72.16
& \textbf{76.78} & 75.82 & 73.37 & 73.43
& \underline{76.21} \\

\midrule
\multicolumn{10}{c}{\textit{Cross-Region}} \\
\midrule

United States
& 61.77 & 61.97 & 59.01 & 55.31
& 60.12 & 70.52 & \underline{79.78} & 78.01
& \textbf{79.98} \\

Germany
& 83.88 & 81.51 & 81.34 & 82.56
& \underline{85.00} & 84.95 & 84.00 & 84.06
& \textbf{85.19} \\

United Kingdom
& 75.67 & 80.55 & 79.14 & 83.36
& 87.64 & 87.82 & \underline{89.19} & 87.13
& \textbf{89.31} \\

France
& 85.13 & 88.43 & 88.00 & 93.33
& 93.24 & 93.98 & 93.80 & \underline{94.97}
& \textbf{95.54} \\

India
& 76.25 & 74.96 & 73.63 & 80.11
& \underline{83.19} & 80.37 & 80.18 & 80.36
& \textbf{84.65} \\

Australia
& 71.12 & 53.76 & 64.01 & 73.21
& 70.74 & 69.76 & 66.85 & \underline{73.53}
& \textbf{77.94} \\

Argentina
& 78.04 & 62.32 & 61.98 & 73.42
& 75.09 & 80.56 & 78.12 & \underline{84.50}
& \textbf{88.15} \\

Brazil
& 74.01 & 58.39 & 69.00 & 76.80
& 86.98 & \textbf{89.03} & \underline{88.62} & 76.82
& 88.40 \\

Vietnam
& 83.70 & 83.04 & 82.62 & 83.76
& 83.53 & 84.06 & 83.21 & \underline{84.27}
& \textbf{84.89} \\

Cambodia
& 74.31 & 75.32 & 63.80 & 63.23
& 70.61 & \underline{75.72} & 73.06 & 75.01
& \textbf{77.18} \\

Canada
& 84.65 & 82.59 & 81.38 & 86.30
& 89.76 & 90.81 & \underline{91.06} & 91.04
& \textbf{91.45} \\

\bottomrule
\end{tabular}
}
\caption{In-region and cross-region performance in terms of IoU (\%). The best and second-best results are highlighted in bold and underlined, respectively.}
\label{tab1}
\end{table*}

\subsection{Segmentation Robustness under Spatio-Temporal Ambiguity}

\begin{figure}[t]
\centering
\includegraphics[width=0.9\columnwidth]{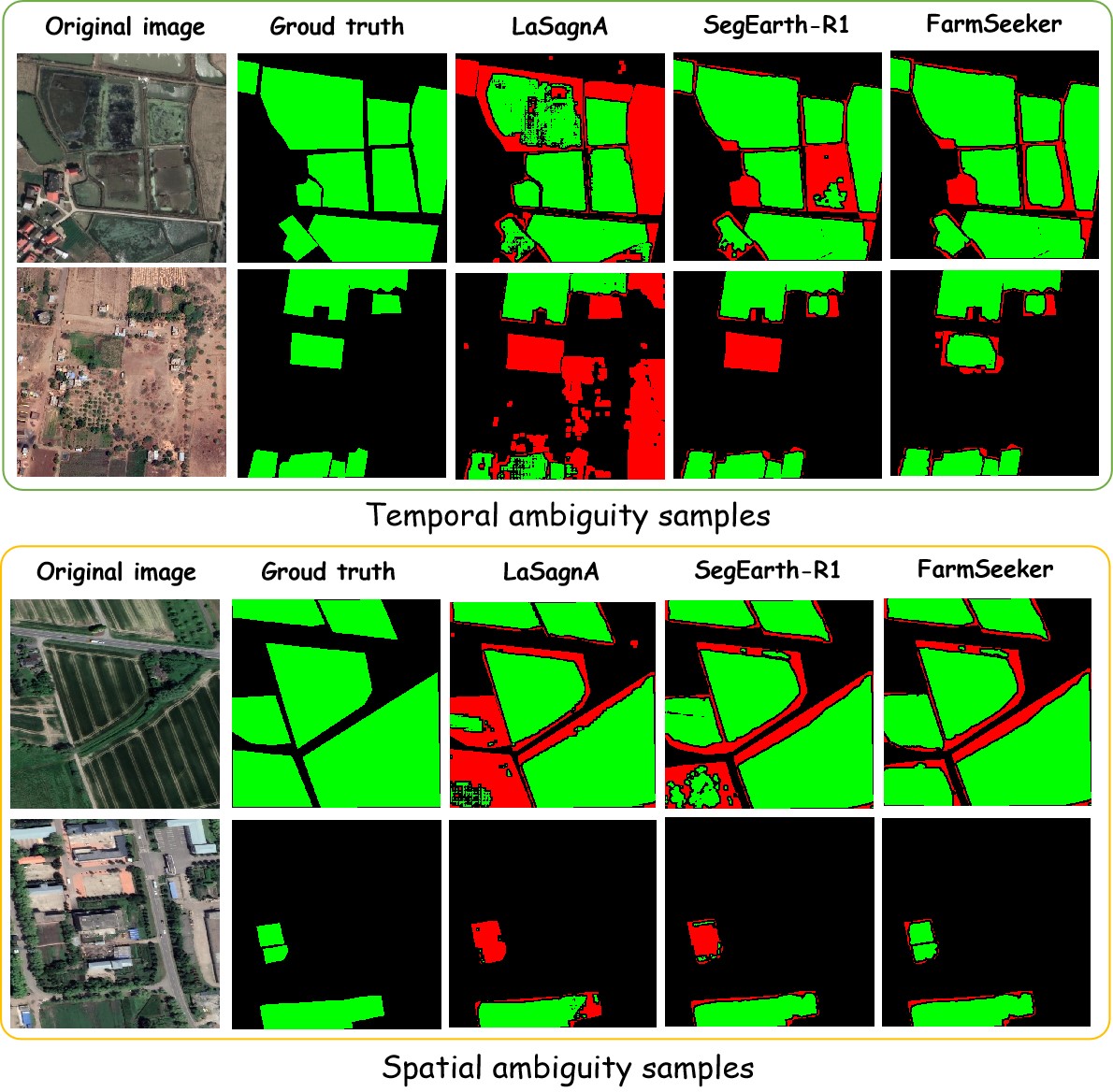} 
\caption{Segmentation results under spatio-temporal ambiguity. Green indicates correctly segmented farmland regions, while red indicates incorrectly segmented regions.}
\label{Exp1(c)}
\end{figure}

To evaluate FarmSeeker under insufficient visual observations, we select 649 highly ambiguous samples from GSFS-Bench, covering temporal ambiguity caused by phenological and seasonal variations, and spatial ambiguity caused by limited fields of view and missing neighborhood context. As shown in Fig.~\ref{Exp1(c)}, the comparison between FarmSeeker and the two best-performing baseline methods shows that FarmSeeker acquires spatio-temporal evidence according to the source of ambiguity, thereby producing more complete and accurate segmentation results. Overall, these results validate the effectiveness of extra spatio-temporal evidence in alleviating segmentation ambiguity under insufficient observations.

\subsection{The Necessity of On-Demand Extra Spatio-Temporal Information Gain}
As shown in Table~\ref{tab:closed_loop_information_gain}, both enhancing local evidence within the current image and incorporating extra-image spatio-temporal observations improve segmentation performance, with more pronounced gains in highly ambiguous regions. This indicates that information gain can effectively alleviate ambiguity in farmland segmentation. However, more observations do not necessarily lead to better performance. The predefined ST strategy directly combines all spatial and temporal observations, achieving 87.54\% Recall and 83.24\% IoU on the China test set, yet underperforming MT, which uses only multi-temporal observations. This suggests that indiscriminate information input may introduce redundant or mismatched evidence. In contrast, FarmSeeker selects spatio-temporal observations on demand according to the information gaps in the current prediction and achieves the best performance. These results demonstrate that its advantage does not come from simply increasing the number of extra images, but from acquiring spatio-temporal evidence that matches the current information gaps on demand.

\begin{table}[t]
\centering
\resizebox{\columnwidth}{!}{%
\begin{tabular}{@{}llcccc@{}}
\toprule
\multirow{2}{*}{Scope / Acquisition}
& \multirow{2}{*}{Setting}
& \multicolumn{2}{c}{China}
& \multicolumn{2}{c}{Ambiguous} \\
\cmidrule(lr){3-4}
\cmidrule(lr){5-6}
& & Recall $\uparrow$ & IoU $\uparrow$
& Recall $\uparrow$ & IoU $\uparrow$ \\
\midrule

Current
& SegEarth-R1
& 85.67 & 82.13
& 67.34 & 63.95 \\
\midrule

Intra-Image
& $+$ ZI
& 86.73 & 83.06
& 76.66 & 72.14 \\
\midrule

\multirow[c]{3}{*}{Extra-Image (Pre.)}
& $+$ SW
& 86.69 & 82.81
& 76.94 & 72.58 \\

& $+$ MT
& \underline{88.61} & \underline{83.70}
& \underline{78.47} & 75.61 \\

& $+$ ST
& 87.54 & 83.24
& 78.31 & \underline{75.76} \\
\midrule

Extra-Image (On-D.)
& $+$ FS
& \textbf{88.63} & \textbf{84.15}
& \textbf{83.45} & \textbf{78.62} \\
\bottomrule

\end{tabular}%
}
\caption{All variants differ only in evidence acquisition. ZI denotes zooming each ambiguous region to $512 \times 512$; SW retrieves a $512 \times 512$ context window centered on the region; MT retrieves all available multi-temporal images of the same region; ST combines SW and MT; FS denotes FarmSeeker.}
\label{tab:closed_loop_information_gain}
\end{table}

\subsection{Ablation Studies}
\textbf{Ablation Study of Progressive Training.} As shown in Table~\ref{tab2}, the complete progressive training strategy achieves the best performance, with Recall and IoU reaching 88.63\% and 84.15\%, respectively. The combination of General FT and Cold Start FT achieves the second-best results, indicating that remote-sensing domain knowledge and task behavior initialization jointly provide an important foundation for subsequent optimization. Adding RFT further improves Recall and IoU by 1.33 and 0.97 percentage points, respectively, validating its effectiveness in further optimizing the model. These results demonstrate that the three training stages play complementary roles in domain adaptation, task initialization, and dynamic decision optimization.

\begin{table}[t]
\centering
\resizebox{\columnwidth}{!}{
\begin{tabular}{ccccc}
  \toprule
  \cmidrule(lr){1-3}
  \cmidrule(lr){4-5}
  General FT
  & Cold Start FT
  & RFT
  & Recall $\uparrow$
  & IoU $\uparrow$ \\
  \midrule

  Yes & --  & --
  & 85.67 & 82.13 \\

  --  & Yes & --
  & 86.64 & 82.59 \\

  --  & --  & Yes
  & 86.63 & 82.79 \\

  --  & Yes & Yes
  & 87.02 & 82.73 \\

  Yes & Yes & --
  & \underline{87.30} & \underline{83.18} \\

  Yes & --  & Yes
  & 86.68 & 82.12 \\

  Yes & Yes & Yes
  & \textbf{88.63} & \textbf{84.15} \\

  \bottomrule
\end{tabular}
}
\caption{Ablation of the progressive training strategy (\%).}
\label{tab2}
\end{table}

\textbf{Ablation Study of Task-driven Rewards.} As shown in Table~\ref{tab:reward_ablation}, applying either $R_{\mathrm{det}}$ or $R_{\mathrm{rea}}$ individually improves the final segmentation performance over the model without RFT. Jointly optimizing both task-driven rewards achieves the best Recall and IoU, demonstrating their complementary contributions to ambiguity perception, tool calling, and collaborative extra-image reasoning.

\begin{table}[t]
\centering
\begin{tabular}{lcc}
\toprule
Setting & Recall $\uparrow$ & IoU $\uparrow$ \\
\midrule
w/o RFT & 87.30 & 83.18 \\
$R_{\mathrm{det}}$ only & 87.64 & 83.65 \\
$R_{\mathrm{rea}}$ only & 87.82 & 83.76 \\
Full rewards & \textbf{88.63} & \textbf{84.15} \\
\bottomrule
\end{tabular}
\caption{Ablation study of the task-driven rewards (\%). }
\label{tab:reward_ablation}
\end{table}

\subsection{Framework Generality and Practical Boundary}
\begin{table}[t]
\centering
\resizebox{.95\columnwidth}{!}{
\begin{tabular}{lcccc}
  \toprule
  Method
  & \multicolumn{2}{c}{Recall}
  & \multicolumn{2}{c}{IoU} \\
  \cmidrule(lr){2-3}
  \cmidrule(lr){4-5}
  & Base & Corrected
  & Base & Corrected \\
  \midrule

  LaSagnA
  & 83.69 & 86.64~{\small($\uparrow$2.95)}
  & 78.19 & 80.75~{\small($\uparrow$2.56)} \\

  PixelLM
  & 84.66 & 87.51~{\small($\uparrow$2.85)}
  & 81.13 & 82.21~{\small($\uparrow$1.08)} \\

  SegEarth-R1
  & 85.67 & 88.63~{\small($\uparrow$2.96)}
  & 82.13 & 84.15~{\small($\uparrow$2.02)} \\

  FSVLM
  & 84.36 & 87.75~{\small($\uparrow$3.39)}
  & 80.98 & 82.13~{\small($\uparrow$1.15)} \\

  \bottomrule
\end{tabular}
}
\caption{Framework generality analysis (\%). Base denotes the initial segmentation result, while Corrected denotes the result refined by FarmSeeker.}
\label{tab5}
\end{table}

To examine whether FarmSeeker depends on a specific base segmentation model, we compare the segmentation performance of different models used as the basic perception tool within FarmSeeker. As shown in Table 5, FarmSeeker consistently improves performance across all four basic perception tools. Recall increases by 2.85-3.39 percentage points, while IoU improves by 1.08-2.56 percentage points, indicating that its error-correction capability is not tied to a particular model. 

\begin{table}[!t]
\centering
\setlength{\tabcolsep}{4.5pt}
\resizebox{\columnwidth}{!}{
\begin{tabular}{ccc}
\toprule
\textbf{Ambiguity Perception}
& \textbf{Tool-Calling}
& \textbf{Collaborative Reasoning} \\
\textbf{F1@0.5 (\%)}
& \textbf{Accuracy (\%)}
& \textbf{Accuracy (\%)} \\
\midrule
60.23 & 90.79 & 88.72 \\
\bottomrule
\end{tabular}
}
\caption{Evaluation of the RE in ambiguity perception, tool calling, and collaborative reasoning. }
\label{tab:agent_capabilities}
\end{table}

We further evaluate the intermediate capabilities of the RE. As shown in Table~\ref{tab:agent_capabilities}, it achieves an F1-score of 60.23\% for ambiguity perception at an IoU threshold of 0.5, a tool-calling accuracy of 90.79\%, and a collaborative reasoning accuracy of 88.72\%. These results indicate that the reasoning engine can effectively select appropriate query tools and reason over the acquired extra images. However, ambiguity perception remains relatively challenging, and localization errors at this stage may propagate to subsequent querying and mask refinement. Table~\ref{tab:computational_cost} reports the segmentation performance and inference time under different evidence acquisition strategies. Compared with SegEarth-R1, FarmSeeker incurs higher computational costs because it additionally performs ambiguity perception, image querying, collaborative reasoning, and local mask refinement. However, this additional computation brings more substantial reliability gains in ambiguous scenarios. Moreover, FarmSeeker explicitly presents ambiguous regions, evidence requirements, query results, and refinement rationales during segmentation, making the process verifiable. Therefore, FarmSeeker is more suitable for offline farmland mapping, difficult-region verification, and practical applications that are sensitive to omission errors. Further reducing its computational overhead remains an important direction for future work.

\begin{table}[!t]
\centering
\small
\setlength{\tabcolsep}{4.5pt}
\resizebox{\columnwidth}{!}{
\begin{tabular}{lccccc}
\toprule
\multirow{2}{*}{Setting}
& \multicolumn{2}{c}{China}
& \multicolumn{2}{c}{Ambiguous}
& \multirow{2}{*}{Time (s) $\downarrow$} \\
\cmidrule(lr){2-3}
\cmidrule(lr){4-5}
& Recall $\uparrow$ & IoU $\uparrow$
& Recall $\uparrow$ & IoU $\uparrow$ & \\
\midrule
SegEarth-R1
& 85.67 & 82.13
& 67.34 & 63.95
& \textbf{0.3} \\

Predefined MT
& \underline{88.61} & \underline{83.70}
& \underline{78.47} & \underline{75.61}
& 24.1 \\

Predefined SW
& 86.69 & 82.81
& 76.94 & 72.58
& \underline{23.8} \\

FarmSeeker
& \textbf{88.63} & \textbf{84.15}
& \textbf{83.45} & \textbf{78.62}
& 23.9 \\
\bottomrule
\end{tabular}}
\caption{Performance and computational cost under different strategies. The inference time is the average processing time per image.}
\label{tab:computational_cost}
\end{table}

\section{Conclusion}
This work revisits farmland segmentation from an information bottleneck perspective and proposes a farmland segmentation paradigm driven by spatio-temporal information gain. Rather than indiscriminately introducing more images, the paradigm identifies information gaps in the current prediction and acquires complementary observations on demand to reduce semantic uncertainty. Building on this paradigm, we develop FarmSeeker, which translates newly acquired information gain into pixel-level improvements through ambiguity perception, on-demand querying, collaborative reasoning, and segmentation refinement. We further establish GSFS-Bench to support unified evaluation across diverse agricultural regions, unseen geographies, and spatio-temporally ambiguous scenarios. Experimental results demonstrate that FarmSeeker consistently improves overall segmentation performance, cross-region generalization, prediction stability, and robustness on challenging samples.

\bibliography{aaai2027}


\clearpage
\clearpage

\setcounter{section}{0}
\setcounter{figure}{0}
\setcounter{table}{0}
\setcounter{equation}{0}

\renewcommand{\thesection}{S\arabic{section}}
\renewcommand{\thefigure}{S\arabic{figure}}
\renewcommand{\thetable}{S\arabic{table}}
\renewcommand{\theequation}{S\arabic{equation}}

\twocolumn[
\begin{center}
{\LARGE\bfseries Supplementary Material}
\end{center}
\vspace{1em}
]

\section{1. Overview}
This supplementary material extends the main paper from five aspects,
including data, implementation, experiments, and process visualization, to further improve the completeness of the task setting and the reproducibility of the method. The contents are organized as follows:

\begin{itemize}
    \item \textbf{Section 2} provides detailed statistics of GSFS-Bench
    and FM-Seg69K, including their scale, data distribution, and
    complex-scene characteristics, to illustrate the representativeness,
    complexity, and challenges of the task setting.

    \item \textbf{Section 3} provides detailed implementation configurations, including the query-and-cropping tool, collaborative reasoning and local mask refinement, progressive training settings, the end-to-end inference procedure, and the LLM judge used during reinforcement fine-tuning.

    \item \textbf{Section 4} provides more comprehensive experimental results and analyses. For experiments already reported in the main paper, we supplement additional evaluation metrics and qualitative results. We further compare FarmSeeker with non-expert human interpretation and conduct large-scale mapping experiments to assess its effectiveness and practical potential in complex farmland scenarios.

    \item \textbf{Section 5} presents step-by-step visualizations of the
    dynamic querying and reasoning process of FarmSeeker in complex
    farmland scenarios, providing a more intuitive understanding of its
    working mechanism.
    
\end{itemize}

\section{2. Dataset and Benchmark Details}
\subsection{2.1. GSFS-Bench}
In this section, we provide additional data analyses of GSFS-Bench and details on the construction of its evaluation subsets.

\textbf{(1) Statistical Analysis of the GSFS-Bench}

To further characterize its data composition and representativeness as a dynamic farmland interpretation benchmark, we analyze GSFS-Bench from three aspects: climate type, field morphology, and seasonal distribution.

\begin{figure*}[!t]
\centering
\includegraphics[width=0.9\textwidth]{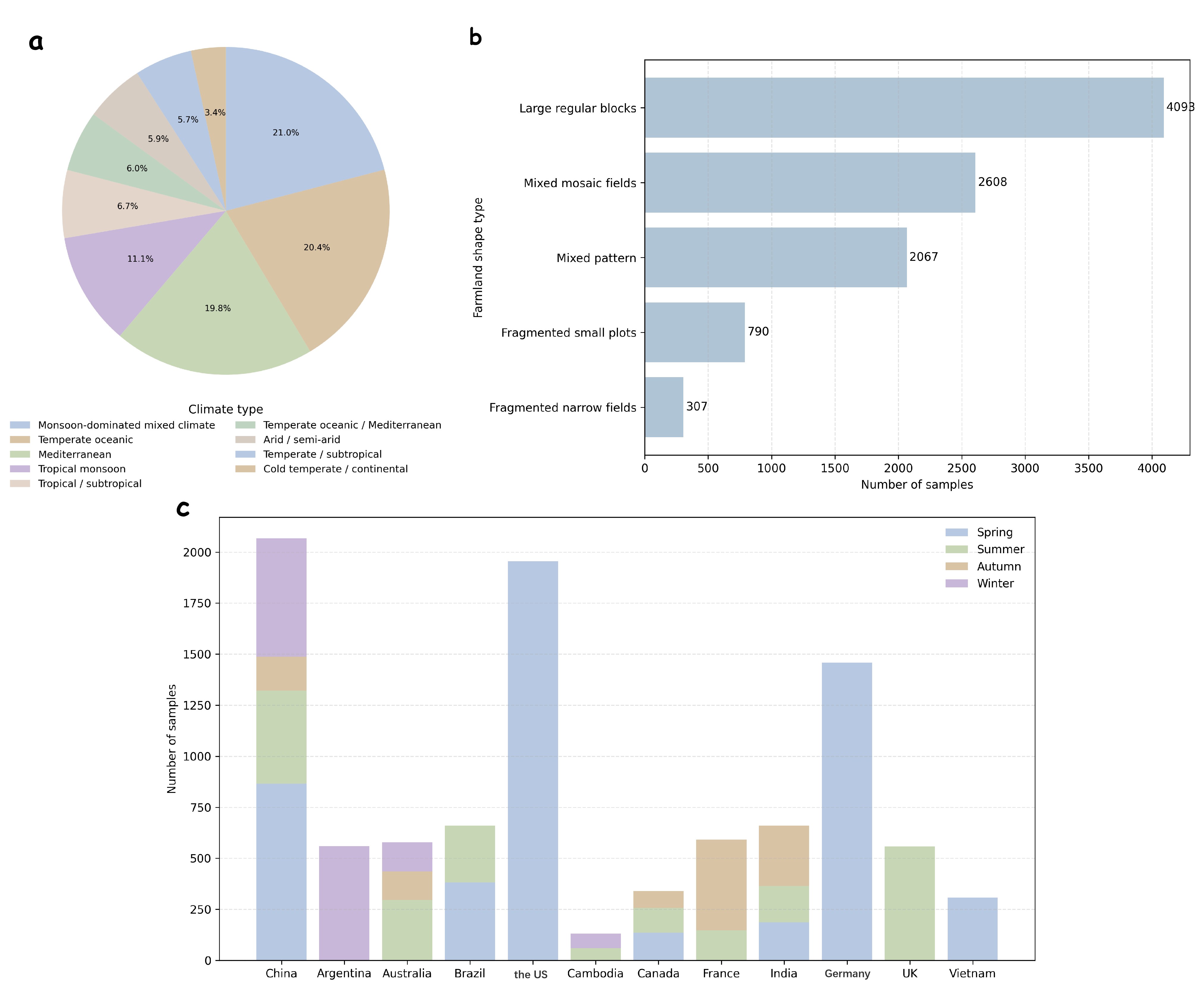} 
\caption{Statistical analysis of GSFS-Bench. 
    (a) Climate-type distribution. 
    (b) Farmland morphology distribution. 
    (c) Seasonal distribution across countries and regions.}
\label{sfig1}
\end{figure*}

As shown in Fig.~\ref{sfig1}(a), GSFS-Bench covers various representative agricultural climate types, including monsoon-dominated mixed climates, temperate oceanic climates, Mediterranean climates, tropical monsoon climates, tropical and subtropical climates, arid and semi-arid climates, and cold temperate continental climates. These climate conditions correspond to substantially different hydrothermal environments, cropping systems, and surface characteristics, resulting in considerable variations in the spectral appearance, boundary clarity, and seasonal evolution of farmland. Therefore, the benchmark provides diverse observation scenarios ranging from regular temperate farmland to fragmented tropical smallholder fields. As shown in Fig.~\ref{sfig1}(b), GSFS-Bench also exhibits substantial diversity in field morphology. Large regular blocks contain 4,093 samples, while Mixed mosaic fields and Mixed pattern contain 2,608 and 2,067 samples, respectively. Fragmented small plots and Fragmented narrow fields contain 790 and 307 samples, respectively. This distribution covers both large-scale regular contiguous farmland and fragmented, narrow, and mosaic-like fields, supporting the evaluation of model adaptability to complex boundaries, small fragmented fields, and heterogeneous backgrounds. As shown in Fig.~\ref{sfig1}(c), samples from different countries and regions exhibit clear seasonal differences. Samples from China cover spring, summer, autumn, and winter, providing relatively complete temporal support for phenological evolution and seasonal ambiguity analysis. Samples from the US and Germany are mainly concentrated in spring, those from Argentina in winter, and those from the UK in summer, while Australia, France, and India cover multiple seasons. This distribution preserves appearance variations across different growth stages, climatic conditions, and agricultural cycles, providing a data foundation for temporal ambiguity analysis and cross-temporal reasoning evaluation.

\textbf{(2) Construction of Specialized Evaluation Subsets in GSFS-Bench}

\textbf{Ambiguity Perception Subset.} We construct reference ambiguous regions from the complete China subset using a multi-base-model voting mechanism. Specifically, multiple farmland segmentation models are first employed to generate prediction masks, and connected regions jointly mis-segmented by at least three models are identified as candidate ambiguous regions. The candidate regions are then filtered according to their bounding-box areas and the proportions of erroneous pixels within the boxes. The bounding-box area is restricted to 1,000--100,000 pixels, and the minimum erroneous-pixel ratio is set to 30\%.
\textbf{Tool-Calling Type Subset.} For each reference ambiguous region, we jointly examine the current image, multi-temporal images, and wide-swath spatial images. Qwen-VL-Max is first employed to generate the required evidence-type label, including temporal, spatial, or none. The generated labels are subsequently manually verified and corrected to ensure the accuracy of the evidence-need annotations. A sample containing multiple ambiguous regions may involve both temporal and spatial queries.
\textbf{Multi-Image Collaborative Reasoning Subset.} According to the annotated evidence type, we retrieve additional images from the corresponding temporal and spatial evidence pools. Human annotators then organize the current image, initial segmentation result, target ambiguous region, and corresponding additional evidence into multi-image collaborative reasoning samples. Based on the ground-truth farmland mask, each target region is further labeled as either farmland or non-farmland.

\subsection{2.2. Statistical Analysis of FM-Seg69K}
\begin{figure*}[!t]
\centering
\includegraphics[width=0.9\textwidth]{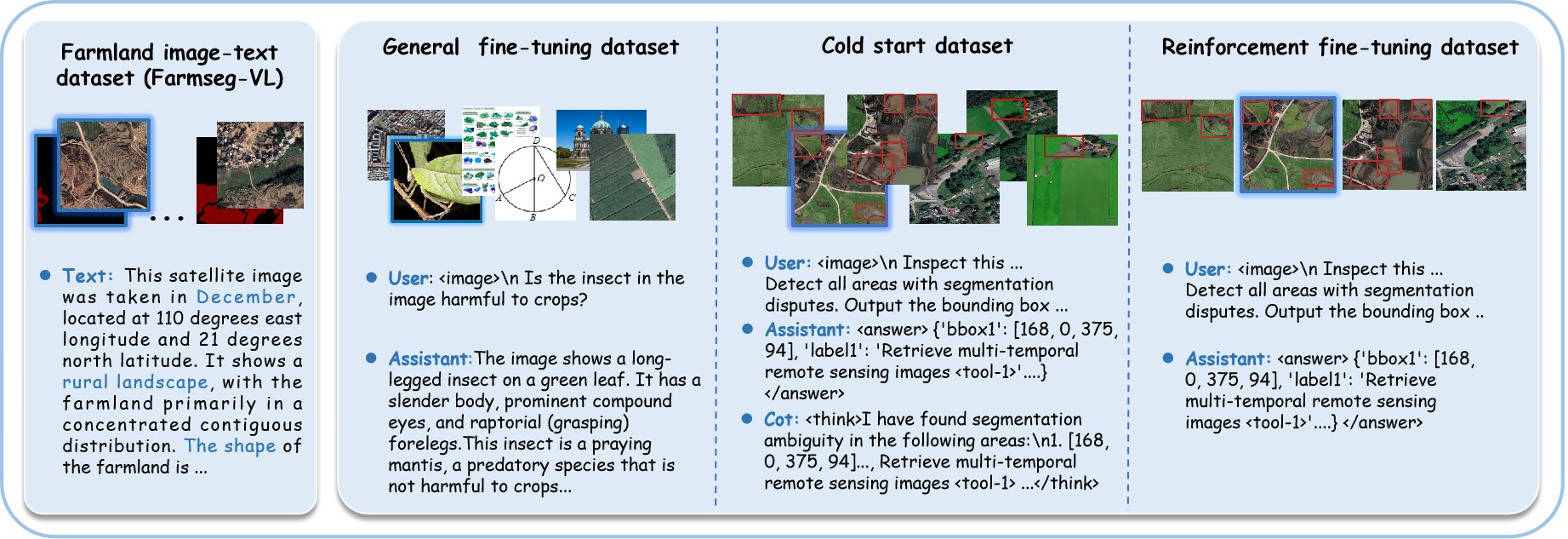} 
\caption{FM-Seg69K sample examples.}
\label{sfig2}
\end{figure*}

\begin{figure}[!t]
\centering
\includegraphics[width=0.9\columnwidth]{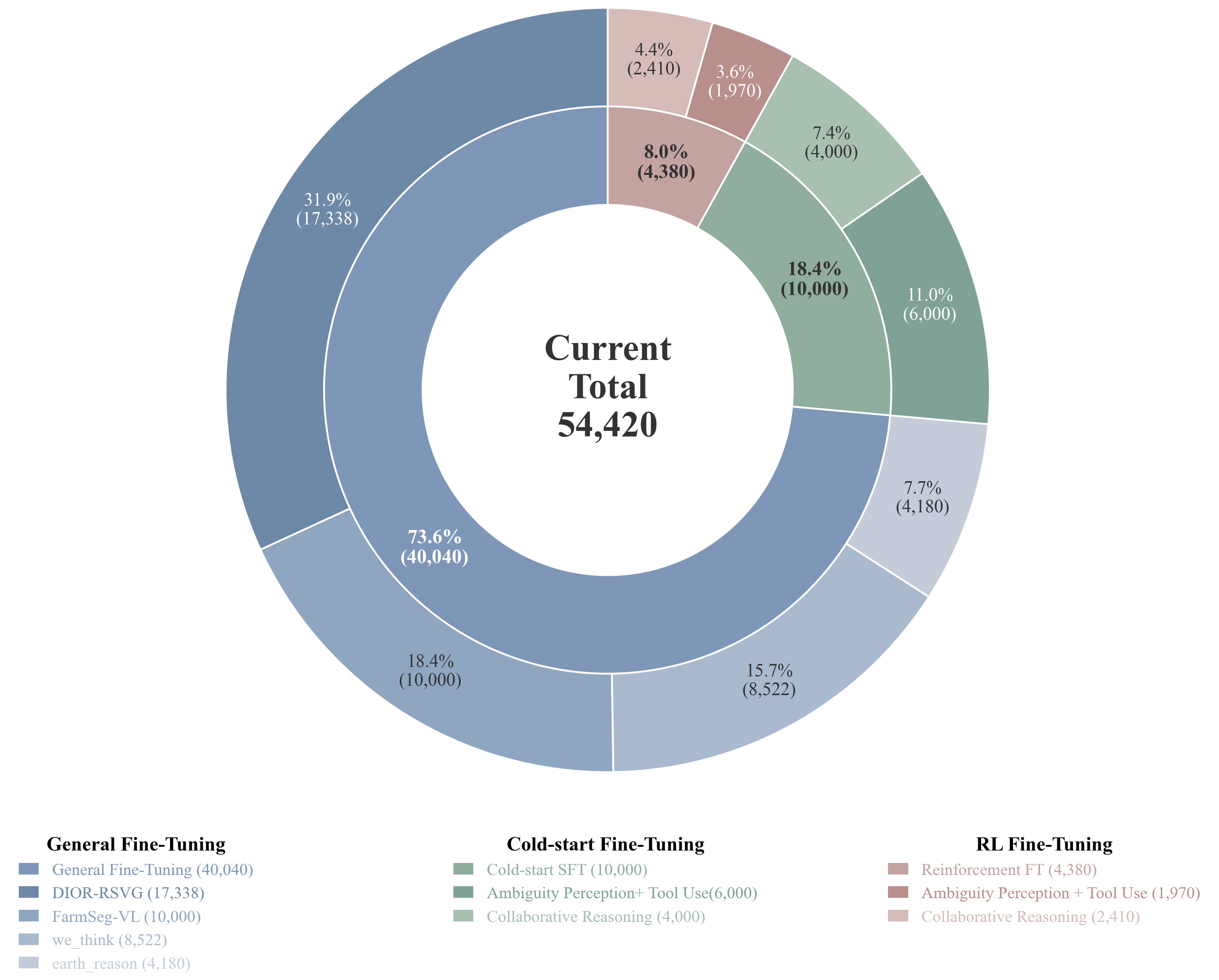}
\caption{Hierarchical composition of FM-Seg69K. The inner ring shows the overall proportions of the three stages used to train the reasoning engine, including General Fine-Tuning, Cold Start Fine-Tuning, and Reinforcement Fine-Tuning. The outer ring further presents the data sources or functional components within each stage.}
\label{sfig3}
\end{figure}

To support the collaborative training of the basic perception tool and the reasoning engine in FarmSeeker, we construct FM-Seg69K. As shown in Fig.~\ref{sfig2}, the basic perception tool is directly trained on the farmland
image-text dataset FarmSeg-VL to acquire fundamental visual-semantic perception capabilities for farmland scenes. The remaining 54,420 samples are used to train the reasoning engine and cover three stages, namely
General Fine-Tuning (GFT), Cold Start Fine-Tuning (CSFT), and Reinforcement Fine-Tuning (RFT). These samples are collectively referred to as FM-Seg54K. As shown in Fig.~\ref{sfig3}, FM-Seg54K follows a clear progressive structure. GFT contains 40,040 samples, accounting for 73.6\%, and is used to establish remote-sensing domain adaptation, cross-modal understanding, and basic reasoning capabilities. CSFT contains 10,000 samples, accounting for 18.4\%, and initializes ambiguity perception, tool calling, and collaborative reasoning behaviors. RFT contains 4,380 samples, accounting for 8.0\%, and further optimizes decision-making and self-correction capabilities in complex scenarios.

\begin{figure*}[!t]
\centering
\includegraphics[width=0.9\textwidth]{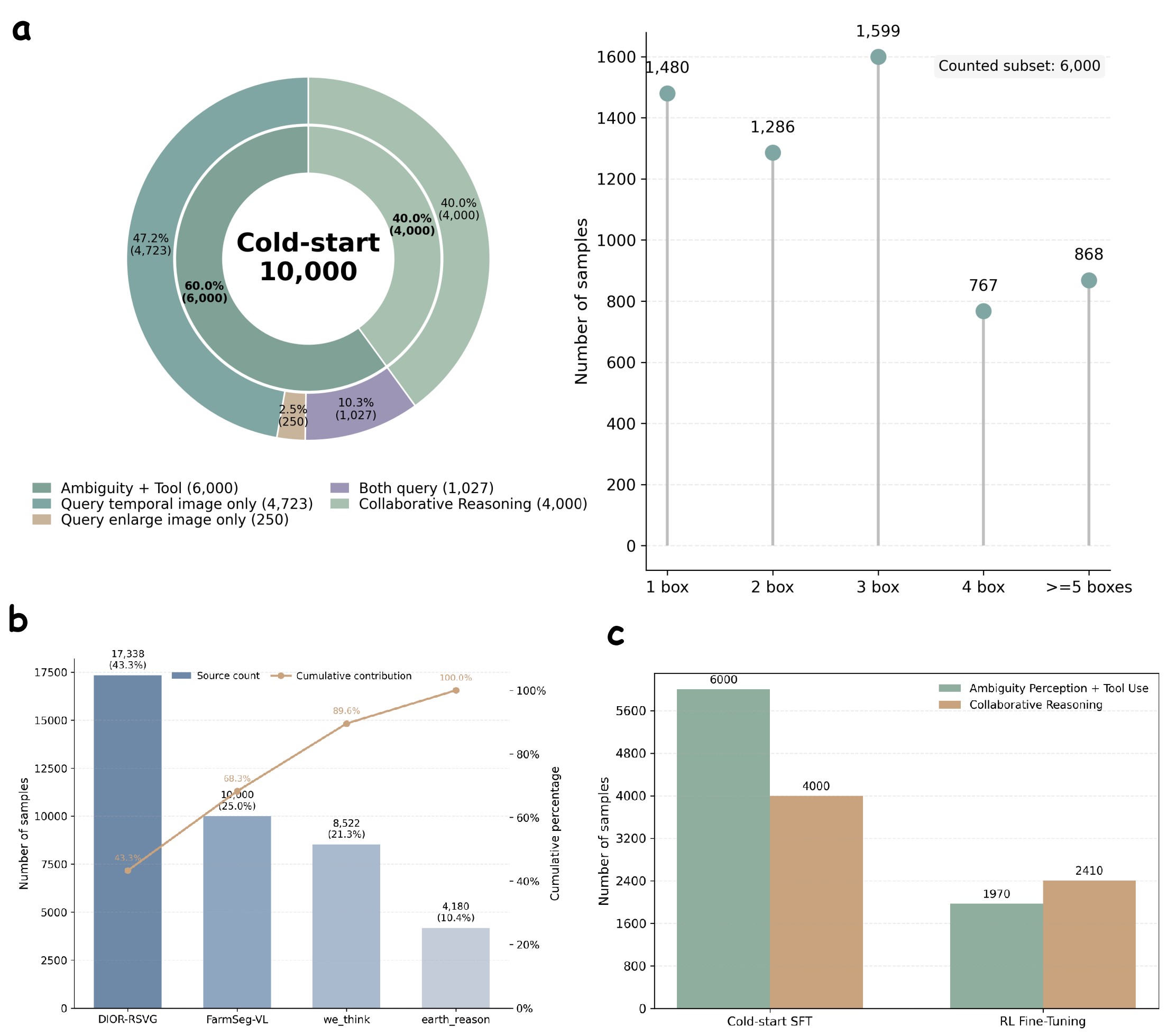} 
\caption{Statistical analysis of FM-Seg54K. (a) Functional composition and difficulty distribution of the Cold Start Fine-Tuning stage. (b) Distribution of data sources in the General Fine-Tuning stage. (c) Comparison of the functional compositions of the Cold Start Fine-Tuning and Reinforcement Fine-Tuning stages.}
\label{sfig4}
\end{figure*}

GFT integrates data for remote-sensing referring segmentation, object detection, general chain-of-thought reasoning, and farmland image-text understanding. As shown in Fig.~\ref{sfig4}(b), DIOR-RSVG, FarmSeg-VL, we\_think, and earth\_reason contain 17,338, 10,000, 8,522, and 4,180 samples, accounting for 43.3\%, 25.0\%, 21.3\%, and 10.4\%, respectively. This stage primarily focuses on remote-sensing scene understanding and farmland semantic supervision, while incorporating general reasoning data to improve the model's transferability and compatibility with reasoning tasks.

CSFT establishes the core task logic of FarmSeeker. As shown in Fig.~\ref{sfig4}(a), this stage contains 6,000 samples for ambiguity perception and tool calling, and 4,000 samples for collaborative reasoning. Among the former, 4,723 samples call the temporal-image query tool, 250 call the larger-context image query tool, and 1,027 call both tools. This distribution indicates that temporal evidence is the primary source of information gain, while complex scenarios still require the joint use of spatial and temporal evidence. This subset also covers different numbers of ambiguous regions. A total of 3,234 samples contain three or more ambiguous regions, accounting for more than half of the subset and enabling the model to learn multi-region inspection, stepwise querying, and cross-evidence reasoning.

RFT follows the same task format as CSFT but places greater emphasis on annotation accuracy, clarity of ambiguity sources, stability of tool mapping, and completeness of reasoning chains. As shown in Fig.~\ref{sfig4}(c), this stage contains 1,970 samples for ambiguity perception and tool calling and 2,410 samples for collaborative reasoning. The proportion of collaborative reasoning samples increases from 40.0\% to 55.0\%, indicating that the training focus shifts from task behavior initialization toward improving reasoning quality and decision-making with the support of external evidence.

\section{3. Implementation Details}
\label{sec:implementation_details}

This section provides implementation details of the query-and-cropping tool, local mask refinement, the end-to-end inference procedure, and the LLM judge used during reinforcement fine-tuning.

\subsection{3.1. Query-and-Cropping Tool}
\label{sec:query_tool}

The query-and-cropping tool $T_q$ maps each ambiguity region $b_i$ predicted by the reasoning engine to corresponding extra-image evidence $E_i$. According to the query strategy $q_i$, it performs either a temporal query or a spatial query.

\textbf{Temporal query.} The tool first matches the current image $X$ to its corresponding wide-swath image and retrieves wide-swath images of the same region from other timestamps according to the region and month information encoded in the file names. Using the geospatial transformations provided by \texttt{rasterio}, the geographic extent of $X$ is projected onto each retrieved image. The corresponding region is then cropped and resampled to the same size as $X$. Since the resulting crop and $X$ share the same geographic coverage and pixel grid, the ambiguity region $b_i$ can be directly overlaid on the temporal crop.

\textbf{Spatial query.} The center of $b_i$ is first converted from the pixel coordinates of $X$ to geographic coordinates and then projected onto the corresponding same-period wide-swath image. A $512\times512$ context window centered on this position is subsequently cropped. The target box is redrawn at the center of the retrieved image while preserving its original size, allowing the reasoning engine to jointly observe the target region and its surrounding spatial context.

Both tools return standardized evidence images with the target region explicitly marked. Their implementation is summarized in Algorithm~\ref{alg:query_tool}.

\begin{algorithm}[!t]
\caption{Query and cropping of extra-image evidence}
\label{alg:query_tool}
\begin{algorithmic}[1]
\REQUIRE Current image $X$; ambiguity region $b_i$; query strategy $q_i$; wide-swath image database $\mathcal{D}$
\ENSURE Extra-image evidence $E_i$

\STATE $W_{\mathrm{cur}} \leftarrow
\operatorname{MatchWideSwathImage}(X,\mathcal{D})$

\IF{$q_i=\texttt{temporal}$}
    \STATE $\mathcal{C} \leftarrow
    \operatorname{RetrieveTemporalImages}
    (W_{\mathrm{cur}},\mathcal{D})$
    \STATE $G_X \leftarrow \operatorname{GeoExtent}(X)$
    \STATE $E_i \leftarrow \varnothing$

    \FOR{each temporal wide-swath image $W_t\in\mathcal{C}$}
        \STATE $P_t \leftarrow
        \operatorname{CropByGeoExtent}(W_t,G_X)$
        \STATE $P_t \leftarrow
        \operatorname{Resize}
        \bigl(P_t,\operatorname{Width}(X),\operatorname{Height}(X)\bigr)$
        \STATE $P_t \leftarrow
        \operatorname{DrawBox}(P_t,b_i)$
        \STATE $E_i \leftarrow E_i\cup\{P_t\}$
    \ENDFOR

\ELSIF{$q_i=\texttt{spatial}$}
    \STATE $p_i^{\mathrm{ctr}} \leftarrow
    \operatorname{Center}(b_i)$
    \STATE $g_i^{\mathrm{ctr}} \leftarrow
    \operatorname{PixelToGeo}(X,p_i^{\mathrm{ctr}})$
    \STATE $p_i^{\mathrm{wide}} \leftarrow
    \operatorname{GeoToPixel}
    (W_{\mathrm{cur}},g_i^{\mathrm{ctr}})$
    \STATE $\Omega_i \leftarrow
    \operatorname{CenteredWindow}
    \bigl(p_i^{\mathrm{wide}},
    \operatorname{Width}(X),
    \operatorname{Height}(X)\bigr)$
    \STATE $P_i \leftarrow
    \operatorname{CropWindow}(W_{\mathrm{cur}},\Omega_i)$
    \STATE $\tilde{b}_i \leftarrow
    \operatorname{RecenterBox}
    \bigl(b_i,\operatorname{Center}(P_i)\bigr)$
    \STATE $P_i \leftarrow
    \operatorname{DrawBox}(P_i,\tilde{b}_i)$
    \STATE $E_i \leftarrow \{P_i\}$
\ENDIF

\RETURN $E_i$
\end{algorithmic}
\end{algorithm}

\subsection{3.2. Collaborative Reasoning and Mask Refinement}
\label{sec:mask_refinement}

After obtaining $E_i$, the reasoning engine $R$ jointly analyzes $X$, the initial mask $M_0$, the ambiguity region $b_i$, and the retrieved evidence. It then generates a regional-level refinement decision $s_i$, which specifies the semantic judgment of the target region and whether local mask correction is required.

For a region requiring correction, the segmentation refinement tool $T_s$ uses $b_i$ as the bounding-box prompt and generates a local refined mask. The refined mask replaces the corresponding prediction in $M_0$, while regions that do not require correction remain unchanged.

When multiple corrected regions overlap, FarmSeeker retains the correction associated with the more confident regional decision. This local replacement strategy converts regional collaborative reasoning into pixel-level refinement without rerunning segmentation over the entire image. The procedure is summarized in Algorithm~\ref{alg:mask_refinement}.

\begin{algorithm}[!t]
\caption{Local mask refinement and fusion}
\label{alg:mask_refinement}
\begin{algorithmic}[1]
\REQUIRE Current image $X$; initial mask $M_0$; regional decisions
$\mathcal{S}=\{(b_i,s_i)\}_{i=1}^{N}$; segmentation refinement tool $T_s$
\ENSURE Final segmentation result $M^*$

\STATE $M^* \leftarrow M_0$

\FOR{$i=1$ to $N$}
    \IF{$s_i$ indicates that no local correction is required}
        \STATE \textbf{continue}
    \ENDIF

    \STATE $M_i \leftarrow T_s(X,b_i,s_i)$
    \COMMENT{Generate a local refined mask using $b_i$ as the prompt}

    \STATE $M^* \leftarrow
    \operatorname{Fuse}(M^*,M_i,b_i,s_i)$
    \COMMENT{Replace the local prediction and resolve possible overlaps}
\ENDFOR

\RETURN $M^*$
\end{algorithmic}
\end{algorithm}

\subsection{3.3. Training Configuration}

Following the progressive training strategy described in the main paper, the reasoning engine is sequentially optimized through General Fine-Tuning (GFT), Cold Start Fine-Tuning (CSFT), and Reinforcement Fine-Tuning (RFT).

\textbf{General and Cold Start Fine-Tuning.} GFT and CSFT are both conducted using LoRA within the LLaMA-Factory framework. GFT establishes general remote-sensing understanding, region localization, and reasoning capabilities, whereas CSFT initializes the task-specific behaviors of ambiguity perception, evidence-need assessment, tool calling, and collaborative reasoning. The two stages adopt the same training configuration. Specifically, the initial learning rate is set to $2\times10^{-5}$, and each stage
is trained for 3 epochs. The per-device batch size is set to 2 with 8 gradient accumulation steps. The maximum number of input-image pixels, denoted by \texttt{image\_max\_pixels}, is set to 262,144.

\textbf{Reinforcement Fine-Tuning.} During RFT, the reasoning engine is optimized using Group Relative Policy Optimization (GRPO). The initial learning rate is set to $5\times10^{-6}$, and the model is trained for 1 epoch with a batch size of 1. For each input sample, the model generates six candidate responses, corresponding to a GRPO group size of 6.

During response generation, the sampling temperature and Top-p are set to 1.0 and 0.9, respectively. The maximum completion length is set to 1,024 tokens, and \texttt{image\_max\_pixels} is set to 12,845,056. A larger image-pixel budget is adopted during RFT because collaborative reasoning samples contain multiple images, including the current observation and the retrieved spatio-temporal evidence. The coefficient of the KL-divergence penalty is set to 0.05 to constrain the distributional deviation of the policy model from the reference model.

For tasks involving multiple reward components, the individual reward terms are first linearly summed to obtain the total reward for each candidate response. The total rewards within each group are then scaled and normalized, and the resulting normalized rewards are used to compute the advantages for policy optimization.

\subsection{3.4. End-to-End Inference Procedure}
\label{sec:inference_procedure}

Given an FRSI $X$, the basic perception tool $T_p$ first generates the initial farmland mask $M_0$. The reasoning engine then jointly analyzes $X$ and $M_0$ to identify ambiguity regions $b_i$ and assign each region a query strategy $q_i$, including temporal, spatial, or none.

For a temporal query, $T_q$ retrieves images of the same region from other timestamps. For a spatial query, it retrieves a context window centered on the ambiguity region from the corresponding same-period wide-swath image. When no additional evidence is required, the reasoning engine directly evaluates the region using the current observation and initial prediction.

The reasoning engine subsequently combines the available evidence to produce the regional-level refinement decision $s_i$. Regions requiring correction are processed by $T_s$, and their refined masks are locally fused with $M_0$ to obtain the final mask $M^*$.

Unlike predefined evidence acquisition strategies, FarmSeeker triggers extra-image querying and local refinement only for regions with insufficient observations. Its additional computation is therefore concentrated on ambiguity regions rather than uniformly applied to the entire image. The complete procedure is presented in Algorithm~\ref{alg:inference}.

\begin{algorithm}[!t]
\caption{End-to-end inference procedure of FarmSeeker}
\label{alg:inference}
\begin{algorithmic}[1]
\REQUIRE Current image $X$; reasoning engine $R$; basic perception tool $T_p$; query-and-cropping tool $T_q$; segmentation refinement tool $T_s$
\ENSURE Final segmentation result $M^*$

\STATE $M_0 \leftarrow T_p(X)$
\COMMENT{Generate the initial farmland mask}

\STATE $\mathcal{A} \leftarrow R(X,M_0)$
\COMMENT{$\mathcal{A}=\{(b_i,q_i)\}_{i=1}^{N}$}

\IF{$\mathcal{A}=\varnothing$}
    \RETURN $M_0$
\ENDIF

\STATE $\mathcal{S}\leftarrow\varnothing$

\FOR{each $(b_i,q_i)\in\mathcal{A}$}
    \IF{$q_i=\texttt{none}$}
        \STATE $E_i\leftarrow\varnothing$
    \ELSE
        \STATE $E_i\leftarrow T_q(X,b_i,q_i)$
        \COMMENT{Retrieve temporal or spatial evidence}
    \ENDIF

    \STATE $s_i\leftarrow R(X,M_0,b_i,E_i)$
    \COMMENT{Perform regional collaborative reasoning}

    \STATE $\mathcal{S}\leftarrow
    \mathcal{S}\cup\{(b_i,s_i)\}$
\ENDFOR

\STATE $M^*\leftarrow
\operatorname{LocalMaskRefinement}
(X,M_0,\mathcal{S},T_s)$
\COMMENT{Apply Algorithm~\ref{alg:mask_refinement}}

\RETURN $M^*$
\end{algorithmic}
\end{algorithm}

\subsection{3.5. LLM Judge for Reinforcement Fine-Tuning}
\label{sec:llm_judge}

During reinforcement fine-tuning of collaborative reasoning, Qwen3-32B is used as a fixed LLM judge to evaluate the reasoning trajectory generated by the reasoning engine. The judge model is used only for reward computation and remains frozen throughout training.

For each response, the question, reasoning content, and final answer are inserted into the placeholders \texttt{\{Question\}}, \texttt{\{Model\_thinking\}}, and \texttt{\{Model\_Answer\}}, respectively. The judge evaluates semantic consistency $r_{\mathrm{con}}$, linguistic and logical correctness $r_{\mathrm{gram}}$, and trajectory compliance $r_{\mathrm{traj}}$.

The complete judge prompt is as follows:

\begin{quote}
\small
\ttfamily
'\{Model\_thinking\}' is the thinking content output by the model for the problem `\{Question\}', and `\{Model\_Answer\}' is the answer output by the model. Firstly, check whether the model is thinking according to the requirements of the problem. Secondly, check whether the thinking content is consistent with the output answer. Finally, check whether the thinking content of the model is coherent in word order and logic. If there are no issues with the above three requirements, then it is judged as 1. If any item does not meet the requirements, it will be judged as 0. Only output judgment, do not output any other content.
\end{quote}

The judge output is parsed as the binary reasoning reward
$R_{\mathrm{cot}}$. An exact output of \texttt{1} is assigned
$R_{\mathrm{cot}}=1$, while all other outputs, including outputs
containing additional text, are assigned $R_{\mathrm{cot}}=0$.
The resulting reward is combined with the answer reward and the
reasoning-format reward according to the reward aggregation defined
in the main paper.

\section{4. Additional Experimental Results}
\subsection{4.1. Additional Evaluation Metrics and Qualitative Results for the Reported Experiments}

This section supplements selected evaluation metrics for the experiments reported in the main paper.

\textbf{(1) Supplementary Results for In-Region Performance and Cross-Region Generalization} 

As shown in Tables~\ref{tab:regional_full_results} and \ref{tab:cross_region_full_results}, we supplement the complete evaluation metrics for the In-Region Performance and Cross-Region Generalization experiments reported in the main paper, including IoU, Recall, and Accuracy. These metrics exhibit trends consistent with those reported in the main paper, further confirming the stable advantages of FarmSeeker in both in-region segmentation and cross-region generalization.

\begin{figure*}[!t]
\centering
\includegraphics[width=0.9\textwidth]{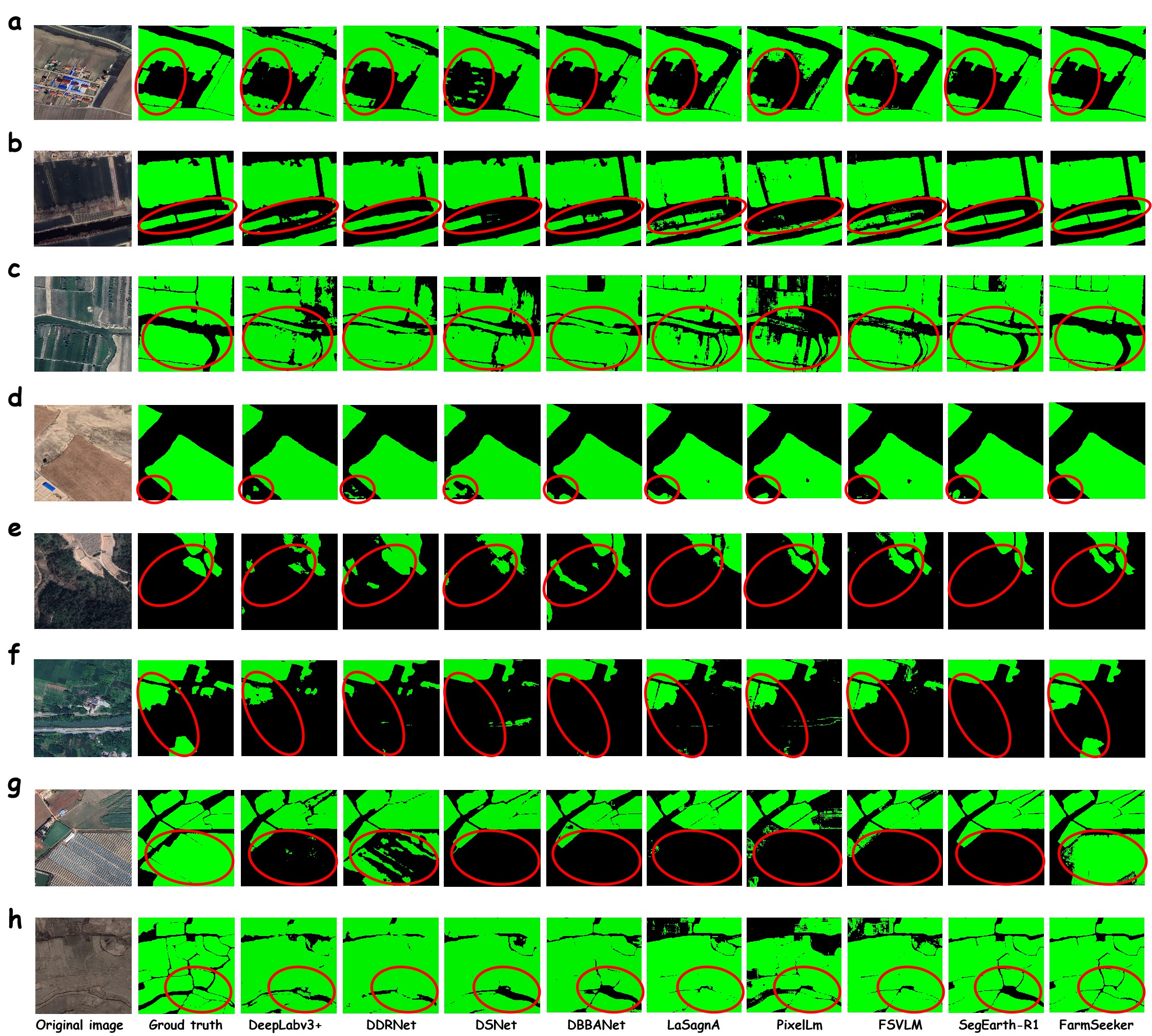} 
\caption{Qualitative comparison of farmland segmentation between FarmSeeker and mainstream methods across different agricultural regions. (a) Selected segmentation results on the Northeast China Plain. (b) Selected segmentation results on the Huang-Huai-Hai Plain. (c) Selected segmentation results in the Northern Arid and Semi-arid Region. (d) Selected segmentation results on the Loess Plateau. (e) Selected segmentation results in South China Areas. (f) Selected segmentation results in the Sichuan Basin. (g) Selected segmentation results on the Yunnan-Guizhou Plateau. (h) Selected segmentation results on the Middle-lower Yangtze Plain. Red circles indicate regions with notable differences. Green masks represent farmland, while black denotes the non-farmland background.}
\label{fig:in_region_qualitative}
\end{figure*}

We further provide qualitative segmentation results of different methods across eight representative agricultural regions in China. As shown in Fig.~\ref{fig:in_region_qualitative}, the qualitative results are generally consistent with the quantitative comparisons. For example, in the densely distributed water-network regions of the Huang-Huai-Hai Plain and Sichuan Basin, as shown in Figs.~\ref{fig:in_region_qualitative}(b) and \ref{fig:in_region_qualitative}(f), the red-circled regions exhibit
substantial confusion between farmland and water bodies or shadows.
Existing methods frequently miss farmland regions or produce excessively expanded masks. In contrast, FarmSeeker acquires complementary spatio-temporal evidence on demand, allowing it to identify ambiguous regions more accurately and recover the boundaries of fragmented fields. Its predictions therefore show higher consistency with the ground-truth annotations. These results further demonstrate that extra spatio-temporal information can effectively alleviate insufficient local observations and semantic ambiguity in complex agricultural scenes.

\begin{table*}[!t]
\centering
\resizebox{\textwidth}{!}{
\begin{tabular}{llccccccccc}
\toprule
Region & Metric
& DeepLabv3+ & DDRNet & DSNet & DBBANet
& LaSagnA & PixelLM & FSVLM & SegEarth-R1 & FarmSeeker \\
\midrule

\multirow{4}{*}{Northeast China Plain}
& Recall & 97.66 & 97.05 & 97.50 & \underline{97.91}
& 96.97 & 97.12 & 95.93 & 97.38 & \textbf{98.00} \\
& IoU & 94.81 & 92.57 & 94.19 & 95.06
& 94.16 & 94.57 & 93.93 & \underline{95.47} & \textbf{95.99} \\
& Acc & 96.04 & 94.17 & 95.54 & \underline{97.02}
& 95.55 & 95.87 & 95.40 & 96.68 & \textbf{97.05} \\
\midrule

\multirow{4}{*}{Huang-Huai-Hai Plain}
& Recall & 97.98 & \underline{98.07} & 97.84 & 98.01
& 96.45 & 97.73 & 96.40 & 97.96 & \textbf{98.40} \\
& IoU & 95.73 & 91.38 & 95.28 & \underline{96.50}
& 94.27 & 95.56 & 94.31 & 96.38 & \textbf{96.78} \\
& Acc & 96.87 & 93.37 & 96.53 & \underline{97.45}
& 95.80 & 96.75 & 95.83 & 97.44 & \textbf{97.72} \\
\midrule

\multirow{4}{*}{Northern Arid and Semi-arid Region}
& Recall & 89.62 & 90.58 & 87.01 & 90.81
& \underline{91.76} & 91.30 & 91.12 & 88.14 & \textbf{92.00} \\
& IoU & 85.90 & 85.91 & 82.63 & 88.24
& \underline{88.58} & 87.68 & 87.53 & 85.64 & \textbf{88.73} \\
& Acc & 90.42 & 90.01 & 88.09 & 92.12
& \underline{92.21} & 91.66 & 91.55 & 91.06 & \textbf{92.93} \\
\midrule

\multirow{4}{*}{Loess Plateau}
& Recall & 97.13 & 93.62 & 96.33 & 94.84
& 89.20 & \underline{97.55} & 97.27 & 96.75 & \textbf{97.70} \\
& IoU & 94.72 & 89.40 & 91.93 & 90.77
& 85.16 & 95.13 & \textbf{95.96} & 94.78 & \underline{95.49} \\
& Acc & 96.61 & 93.06 & 94.71 & 93.97
& 90.27 & 97.04 & \textbf{97.44} & 96.80 & \underline{97.23} \\
\midrule

\multirow{4}{*}{South China Areas}
& Recall & 72.51 & \underline{84.59} & 68.49 & 78.76
& 61.38 & 77.72 & 76.81 & 81.01 & \textbf{84.74} \\
& IoU & 65.14 & 63.87 & 59.17 & 71.15
& 53.57 & 69.36 & 69.39 & \underline{73.01} & \textbf{75.23} \\
& Acc & 92.16 & 90.33 & 90.45 & \underline{93.54}
& 89.25 & 93.06 & 90.76 & 93.42 & \textbf{93.87} \\
\midrule

\multirow{4}{*}{Sichuan Basin}
& Recall & 81.89 & 84.57 & 84.35 & 82.86
& 85.37 & \underline{85.85} & 84.40 & 81.83 & \textbf{86.03} \\
& IoU & 77.23 & 76.64 & 77.15 & 79.17
& 79.25 & \underline{79.86} & 78.57 & 78.30 & \textbf{81.57} \\
& Acc & 92.57 & 91.51 & 92.31 & 93.29
& 93.12 & \underline{93.34} & 92.91 & 93.32 & \textbf{94.27} \\
\midrule

\multirow{4}{*}{Yunnan-Guizhou Plateau}
& Recall & 89.46 & 92.49 & 91.18 & 91.82
& 81.10 & 87.69 & \underline{93.25} & 92.55 & \textbf{94.99} \\
& IoU & 81.20 & 66.21 & 81.46 & 80.74
& 73.96 & 78.76 & \underline{85.03} & 84.73 & \textbf{86.41} \\
& Acc & 91.34 & 79.42 & 91.33 & 93.24
& 88.07 & 90.11 & 92.60 & \underline{93.36} & \textbf{94.05} \\
\midrule

\multirow{4}{*}{Middle-lower Yangtze Plain}
& Recall & 78.68 & 79.76 & 79.21 & 79.87
& 76.60 & \underline{80.12} & 76.92 & 76.19 & \textbf{81.56} \\
& IoU & 75.13 & 75.20 & 74.52 & 72.16
& \textbf{76.78} & 75.82 & 73.37 & 73.43 & \underline{76.21} \\
& Acc & 86.27 & 85.26 & 85.72 & 85.28
& 84.42 & 86.53 & \underline{86.89} & 85.46 & \textbf{87.27} \\

\bottomrule
\end{tabular}
}
\caption{Regional segmentation performance in terms of Recall, IoU, and Accuracy (\%).}
\label{tab:regional_full_results}
\end{table*}

\begin{table*}[!t]
\centering
\resizebox{\textwidth}{!}{
\begin{tabular}{llccccccccc}
\toprule
Country & Metric
& DeepLabv3+ & DDRNet & DSNet & DBBANet
& LaSagnA & PixelLM & FSVLM & SegEarth-R1 & FarmSeeker \\
\midrule

\multirow{4}{*}{United States}
& Recall
& 63.20 & 65.22 & 60.67 & 56.42
& 61.51 & 72.73 & 80.83 & \underline{80.94}
& \textbf{83.04} \\
& IoU
& 61.77 & 61.97 & 59.01 & 55.31
& 60.12 & 70.52 & \underline{79.78} & 78.01
& \textbf{79.98} \\
& Acc
& 67.76 & 67.01 & 65.27 & 62.44
& 66.37 & 74.94 & 82.05 & \underline{82.31}
& \textbf{83.54} \\
\midrule

\multirow{4}{*}{Germany}
& Recall
& 85.20 & 84.01 & 82.62 & 83.81
& \underline{86.96} & 86.02 & 85.46 & 85.22
& \textbf{87.03} \\
& IoU
& 83.88 & 81.51 & 81.34 & 82.56
& \underline{85.00} & 84.95 & 84.00 & 84.06
& \textbf{85.19} \\
& Acc
& 88.42 & 86.52 & 86.59 & 87.47
& 89.15 & \underline{89.22} & 88.48 & 88.57
& \textbf{89.29} \\
\midrule

\multirow{4}{*}{United Kingdom}
& Recall
& 80.30 & 87.54 & 83.11 & 85.49
& 89.28 & \underline{90.26} & 90.16 & 88.87
& \textbf{91.68} \\
& IoU
& 75.67 & 80.55 & 79.14 & 83.36
& 87.64 & 87.82 & \underline{89.19} & 87.13
& \textbf{89.31} \\
& Acc
& 82.53 & 85.70 & 85.18 & 88.46
& 91.49 & 91.53 & \textbf{92.53} & 91.12
& \underline{91.99} \\
\midrule

\multirow{4}{*}{France}
& Recall
& 90.29 & 90.22 & 92.25 & 95.68
& 95.36 & 96.00 & 95.56 & \underline{96.74}
& \textbf{97.17} \\
& IoU
& 85.13 & 88.43 & 88.00 & 93.33
& 93.24 & 93.98 & 93.80 & \underline{94.97}
& \textbf{95.54} \\
& Acc
& 88.03 & 90.44 & 90.46 & 94.81
& 94.75 & 95.33 & 95.21 & \underline{96.11}
& \textbf{97.98} \\
\midrule

\multirow{4}{*}{India}
& Recall
& 79.69 & 80.23 & 77.45 & 85.95
& 87.95 & 86.00 & 87.85 & \underline{88.43}
& \textbf{90.82} \\
& IoU
& 76.25 & 74.96 & 73.63 & 80.11
& \underline{83.19} & 80.37 & 80.18 & 80.36
& \textbf{84.65} \\
& Acc
& 83.06 & 81.71 & 81.07 & 85.54
& 85.18 & 85.00 & 85.18 & \underline{87.95}
& \textbf{88.76} \\
\midrule

\multirow{4}{*}{Australia}
& Recall
& 87.02 & 86.10 & 87.52 & \underline{88.37}
& 84.86 & 81.14 & 80.12 & 86.60
& \textbf{90.03} \\
& IoU
& 71.12 & 53.76 & 64.01 & 73.21
& 70.74 & 69.76 & 66.85 & \underline{73.53}
& \textbf{77.94} \\
& Acc
& 82.29 & 62.90 & 75.35 & 82.88
& 80.34 & 80.21 & 76.79 & \underline{84.38}
& \textbf{87.23} \\
\midrule

\multirow{4}{*}{Argentina}
& Recall
& 81.01 & 70.21 & 66.28 & 76.00
& 82.38 & 85.50 & 86.04 & \underline{88.34}
& \textbf{92.75} \\
& IoU
& 78.04 & 62.32 & 61.98 & 73.42
& 75.09 & 80.56 & 78.12 & \underline{84.50}
& \textbf{88.15} \\
& Acc
& 84.42 & 71.00 & 72.22 & 81.20
& 81.33 & 85.91 & 83.54 & \underline{88.93}
& \textbf{91.48} \\
\midrule

\multirow{4}{*}{Brazil}
& Recall
& 76.26 & 75.10 & 72.83 & 78.72
& 91.50 & 90.86 & \textbf{94.37} & 78.73
& \underline{93.38} \\
& IoU
& 74.01 & 58.39 & 69.00 & 76.80
& 86.98 & \textbf{89.03} & \underline{88.62} & 76.82
& 88.40 \\
& Acc
& 84.21 & 68.44 & 80.71 & 85.98
& 91.75 & \underline{93.18} & 92.85 & 86.00
& \textbf{93.48} \\
\midrule

\multirow{4}{*}{Vietnam}
& Recall
& \underline{91.47} & 89.53 & 89.36 & 89.57
& 90.94 & 90.53 & 88.92 & 90.15
& \textbf{92.04} \\
& IoU
& 83.70 & 83.04 & 82.62 & 83.76
& 83.53 & 84.06 & 83.21 & \underline{84.27}
& \textbf{84.89} \\
& Acc
& 89.86 & 89.02 & 88.94 & 90.12
& 89.79 & 90.23 & 89.78 & \underline{90.42}
& \textbf{91.67} \\
\midrule

\multirow{4}{*}{Cambodia}
& Recall
& 76.00 & 76.76 & 65.57 & 64.23
& 73.47 & 77.32 & 75.25 & \underline{77.56}
& \textbf{82.85} \\
& IoU
& 74.31 & 75.32 & 63.80 & 63.23
& 70.61 & \underline{75.72} & 73.06 & 75.01
& \textbf{77.18} \\
& Acc
& 80.31 & 80.04 & 72.13 & 72.01
& 77.85 & 81.43 & 79.91 & \underline{81.64}
& \textbf{82.48} \\
\midrule

\multirow{4}{*}{Canada}
& Recall
& 93.16 & 88.83 & 86.39 & 92.79
& 96.28 & 98.16 & \underline{98.35} & 98.11
& \textbf{98.70} \\
& IoU
& 84.65 & 82.59 & 81.38 & 86.30
& 89.76 & 90.81 & \underline{91.06} & 91.04
& \textbf{91.45} \\
& Acc
& 86.21 & 84.71 & 83.86 & 87.97
& 91.03 & 91.81 & 92.12 & \underline{92.47}
& \textbf{92.51} \\

\bottomrule
\end{tabular}
}
\caption{Cross-region farmland segmentation performance in terms of
Recall, IoU, and Accuracy (\%).}
\label{tab:cross_region_full_results}
\end{table*}

\textbf{(2) Supplementary Results on Segmentation Robustness under Spatio-Temporal Ambiguity} 

\begin{figure}[!t]
\centering
\includegraphics[width=0.9\columnwidth]{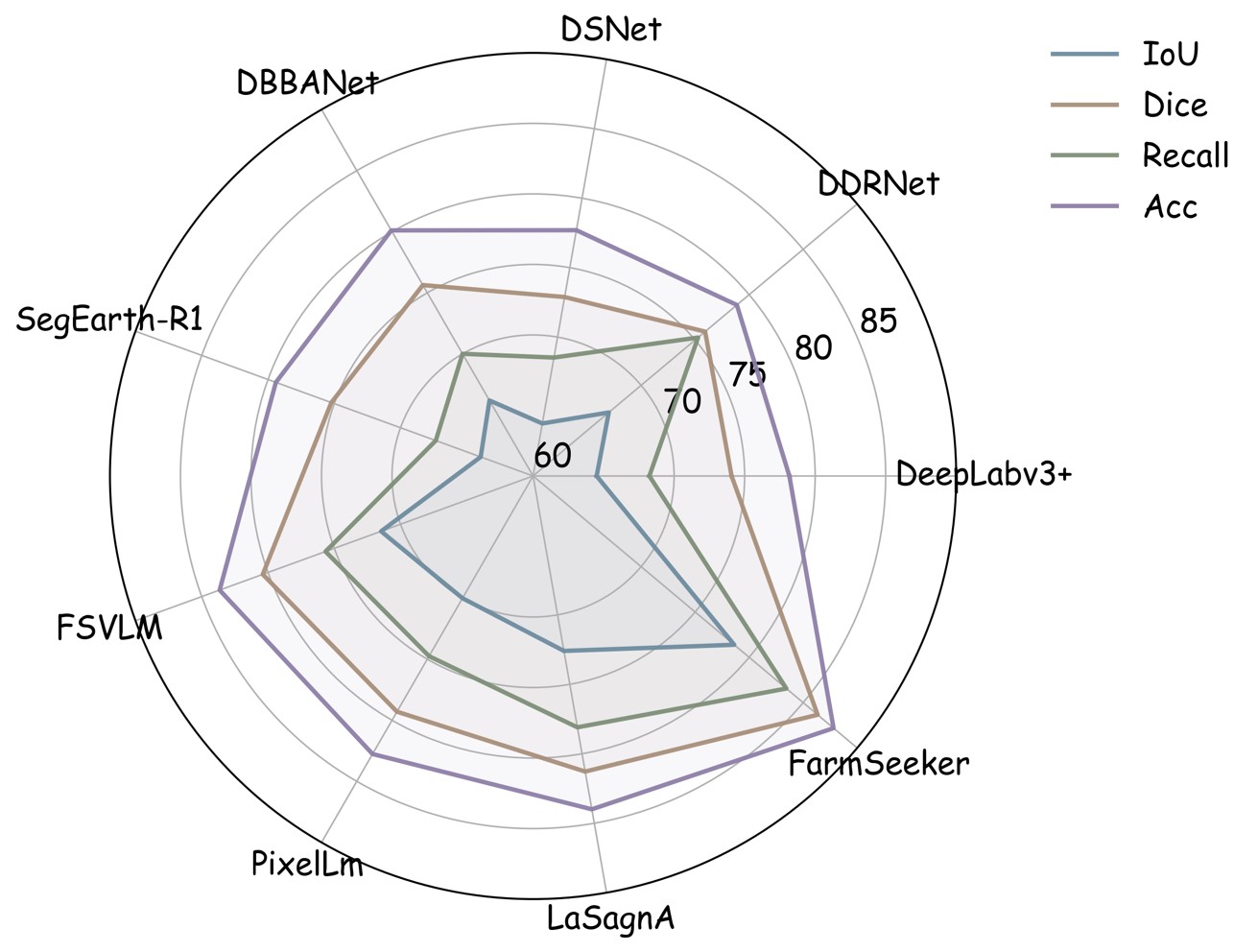} 
\caption{Radar chart comparison of different methods across four metrics, including IoU, Dice, Recall, and Accuracy. The methods are arranged along the axes, while the metrics are represented as closed curves to provide an overall view of each model’s performance profile on the ambiguous sample set.}
\label{ambiguity_robustness}
\end{figure}

This section reports the complete results of the Segmentation Robustness under Spatio-Temporal Ambiguity experiment. As shown in Fig.~\ref{ambiguity_robustness}, FarmSeeker achieves the best
overall performance across all four metrics, including IoU, Dice, Recall,
and Accuracy. SegEarth-R1, which serves as the basic perception tool of
FarmSeeker, exhibits only moderate performance under ambiguous scenarios
and shows a clear gap from FarmSeeker. This indicates that the improvement
of FarmSeeker in highly ambiguous scenarios does not result from simple
enhancement of the basic segmentation output, but primarily from its
active analysis of ambiguous regions, acquisition of complementary
evidence, and targeted refinement.

\begin{figure*}[!t]
\centering
\includegraphics[width=0.9\textwidth]{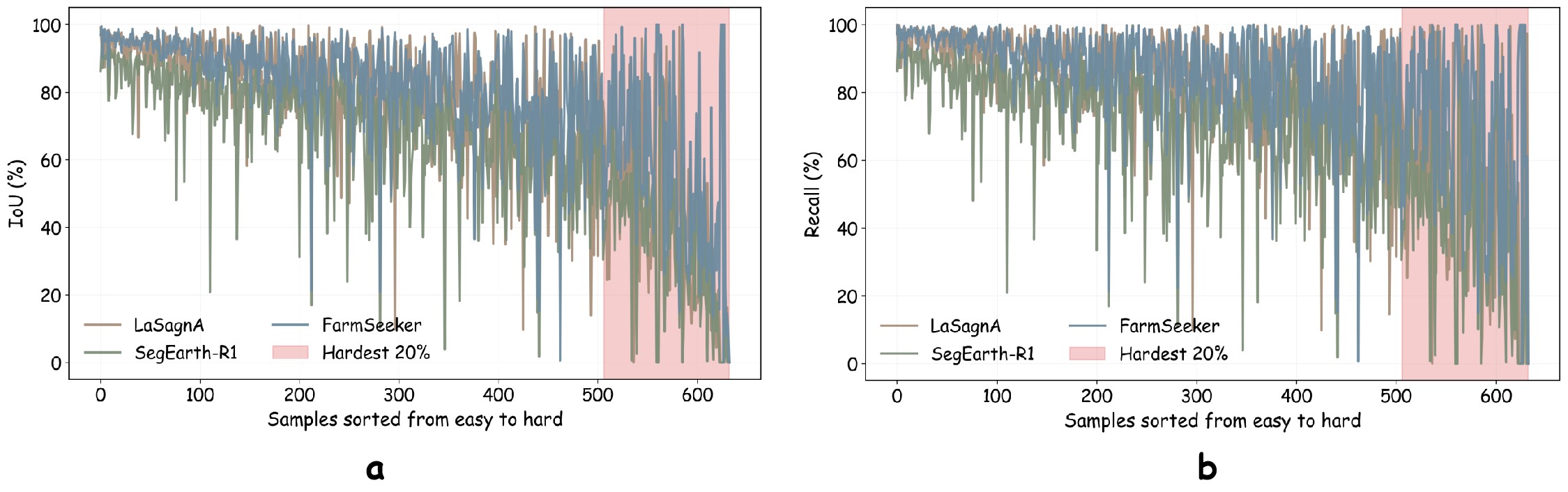} 
\caption{Performance analysis under ambiguous scenarios. (a) Long-tail performance curves based on IoU. Samples are reordered from easy to difficult, where sample difficulty is jointly defined by the mean IoU achieved by all methods on each sample, with a lower mean indicating a more difficult sample. The shaded region denotes the most difficult 20\% of samples. The figure compares the performance trends of the second-best method LaSagnA, the basic segmentation model SegEarth-R1, and the proposed FarmSeeker across the sample distribution from easy to difficult. (b) Long-tail performance curves based on Recall. The same sample ordering and difficult-sample definition as in (a) are adopted to analyze changes in the target recovery capability of different models on long-tail difficult samples.}
\label{ambiguity_robustness2}
\end{figure*}

We further analyze the performance of different methods over the long-tailed distribution of ambiguous samples. As shown in Figs.~\ref{ambiguity_robustness2}(a) and \ref{ambiguity_robustness2}(b), the samples are reordered from easy to
difficult, with the most difficult 20\% highlighted by shaded regions.
For both IoU and Recall, the curve of FarmSeeker remains above those of
LaSagnA and SegEarth-R1 across most of the sample range, indicating that
its advantage is not limited to a small number of easy samples but extends
across the overall distribution of ambiguous samples.

In particular, within the most difficult 20\% of samples, although all
methods exhibit varying degrees of performance degradation, FarmSeeker
shows a relatively slower decline and maintains a higher overall
performance level. This demonstrates that FarmSeeker preserves better
target recovery and regional overlap quality under stronger temporal
variations, more severe boundary confusion, and increasingly complex
spatial structures, reflecting its superior long-tail robustness and
adaptability to complex scenarios.

\textbf{(3) Supplementary Results on Ablation Studies}

This section provides supplementary ablation experiments on different
reasoning engines. As shown in Table~\ref{tabs4}, the general-purpose
Qwen2.5-VL-7B yields the same performance as the setting without a
reasoning engine. This is because the untrained 7B model cannot reliably
generate valid ambiguous regions under the task-specific output format.
Consequently, no extra-image query or subsequent mask refinement is
triggered, and the final prediction remains identical to the output of
the basic perception tool.

As the scale of the general-purpose reasoning engine increases, the
overall system performance generally improves, indicating that stronger
multimodal perception and reasoning capabilities facilitate ambiguity
identification and extra-image evidence integration. However, after
applying the proposed progressive training strategy, the task-trained
Qwen2.5-VL-7B achieves the best overall performance, with Recall, IoU,
and Acc reaching 88.63\%, 84.15\%, and 91.74\%, respectively,
outperforming the substantially larger general-purpose
Qwen2.5-VL-72B. These results demonstrate that the performance gains of
FarmSeeker do not simply rely on increasing model scale. Instead,
progressive task-specific supervision and reinforcement optimization
enable the compact 7B reasoning engine to acquire more effective
ambiguity perception, query planning, and collaborative reasoning
capabilities, thereby surpassing larger general-purpose models.

\begin{table}[t]
\centering
\begin{tabular}{lccc}
  \toprule
  Reasoning Engine
  & Recall $\uparrow$
  & IoU $\uparrow$
  & Acc $\uparrow$ \\
  \midrule
  None
  & 85.67
  & 82.13
  & 90.57 \\
  Qwen2.5-VL-7B
  & 85.67
  & 82.13
  & 90.57 \\
  Qwen2.5-VL-32B
  & 85.94
  & 82.95
  & 90.70 \\
  Qwen2.5-VL-72B
  & \underline{88.37}
  & \underline{83.94}
  & \underline{91.20} \\
  Qwen2.5-VL-7B (T)
  & \textbf{88.63}
  & \textbf{84.15}
  & \textbf{91.74} \\
  \bottomrule
\end{tabular}
\caption{Ablation study of different reasoning engines (\%). 
T denotes the model trained using the proposed progressive training
strategy. The general-purpose Qwen2.5-VL-7B fails to reliably generate
valid ambiguous regions and therefore produces the same result as the
setting without a reasoning engine.}
\label{tabs4}
\end{table}

\textbf{(4) Performance Limits of Ambiguity Perception in FarmSeeker}

\begin{table}[!t]
\centering
\begin{tabular}{lcc}
\toprule
\textbf{Evaluation Metric}
& $\mathbf{IoU}=0.5$
& $\mathbf{IoU}=0.75$ \\
\midrule
Precision (\%) & 63.51 & 22.51 \\
Recall (\%)    & 58.03 & 13.26 \\
F1-score (\%)  & 60.23 & 16.69 \\
\bottomrule
\end{tabular}
\caption{Quantitative results of ambiguity perception under different
IoU matching thresholds.}
\label{tab:ambiguity_perception}
\end{table}

Table~\ref{tab:ambiguity_perception} reports the ambiguity perception results under different IoU matching thresholds. We evaluate the predicted ambiguity boxes using bounding-box-based Precision, Recall, and F1-score at IoU thresholds of 0.5 and 0.75. At an IoU threshold of 0.5, FarmSeeker achieves a Precision of 63.51\%, a Recall of 58.03\%, and an F1-score of 60.23\%. These results indicate that the model has developed a preliminary ability to identify regions with potentially insufficient information from the initial segmentation results. When the IoU threshold increases from 0.5 to 0.75, Precision, Recall, and F1-score decrease by 41.00, 44.77, and 43.54 percentage points, respectively. This suggests that the current model is more effective at coarse-grained localization of ambiguous regions, namely identifying where potential problems may exist, but remains limited in accurately fitting their boundaries. This result also reflects the fundamental difference between ambiguity perception and conventional object detection. Ambiguous regions often lack stable, clear, and structured boundaries, while their spatial extents are inherently fuzzy, transitional, and semantically dependent. A substantial performance decrease under a stricter IoU threshold is therefore expected. In addition, Precision is consistently higher than Recall at both IoU thresholds, indicating that the model adopts a relatively conservative prediction strategy. Although a considerable proportion of the predicted boxes are valid, the coverage of ground-truth ambiguous regions remains incomplete, with missed detections being more prominent. This observation is consistent with the characteristics of complex farmland scenes. Farmland often exhibits substantial visual confusion with bare soil, roads, abandoned fields, and the boundaries of built-up areas. Consequently, the model is more likely to generate localized and conservative candidate regions than to completely cover all ambiguous areas.

Overall, FarmSeeker has demonstrated an initial capability for ambiguity-region perception, providing effective candidate regions for subsequent query decisions and collaborative reasoning. Nevertheless, precise ambiguity localization remains an important challenge for the current framework. At present, the model is more capable of answering where potential problems may exist than where the precise boundaries of the problematic regions lie. Future work may further improve the representation of ambiguous regions through finer-grained region modeling, stronger boundary constraints, and multi-scale contextual modeling.

\subsection{4.2.  A Large-scale Farmland Mapping Applications of FarmSeeker}
To evaluate the potential of FarmSeeker for large-scale farmland mapping, we conduct experiments in a representative region with complex terrain in Sichuan Province, China. The region is characterized by fragmented terrain, small and sparsely distributed farmland parcels, and interwoven land-cover types such as forests, settlements, and water bodies, posing substantial challenges to the recognition of small fields and the preservation of complex boundaries.

\begin{figure*}[!t]
\centering
\includegraphics[width=0.9\textwidth]{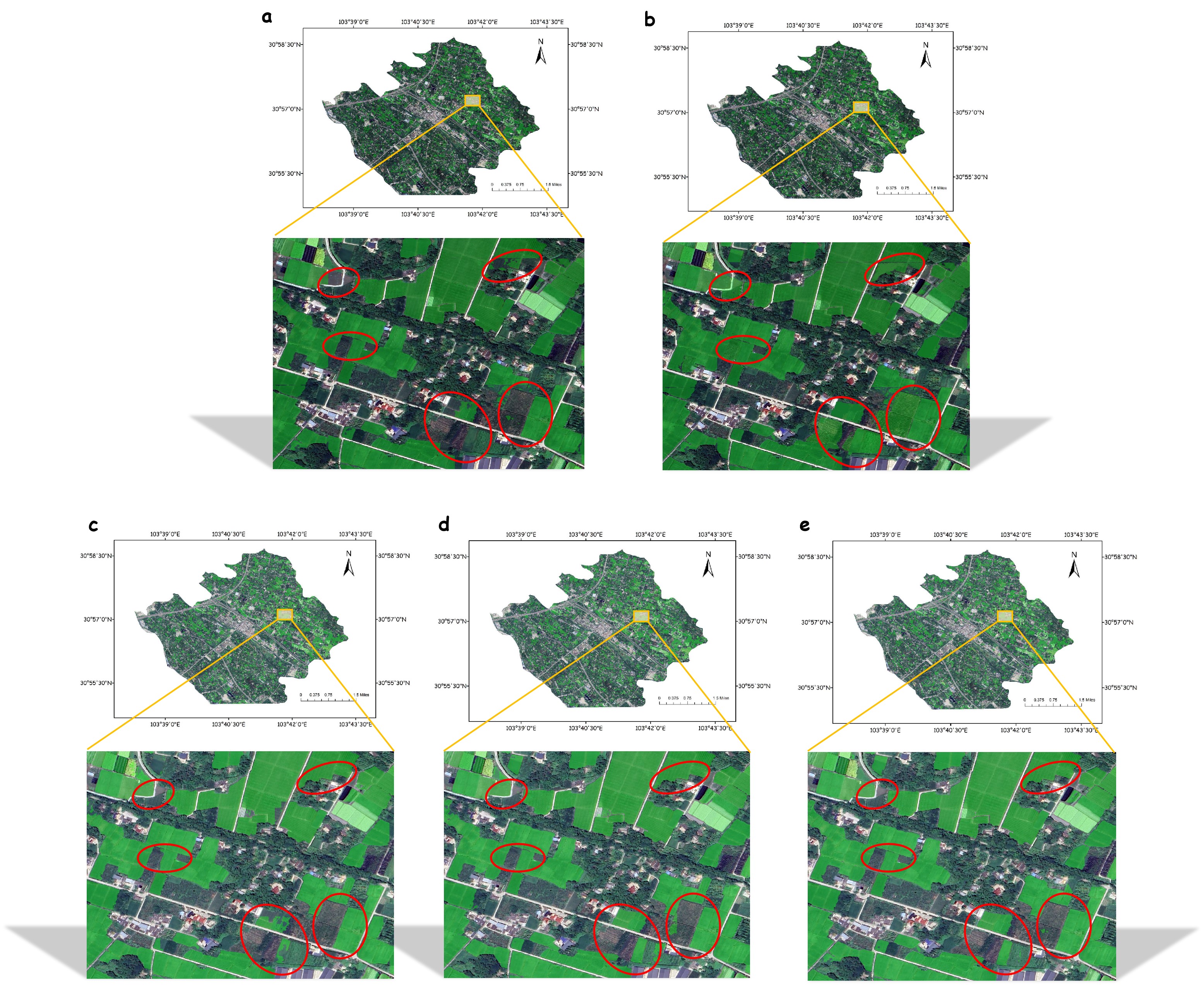} 
\caption{Comparison between FarmSeeker and the sliding-window strategy in a mapping case from Sichuan. a, SegEarth-R1. b, FarmSeeker. c, SegEarth-R1 with sliding-window inference (window size: 512 × 512, overlap ratio: 10\%). d, SegEarth-R1 with sliding-window inference (window size: 512 × 512, overlap ratio: 30\%). e, SegEarth-R1 with sliding-window inference (window size: 512 × 512, overlap ratio: 50\%). The red circles highlight representative ambiguous regions.}
\label{fig:mapping}
\end{figure*}

As shown in Fig.~\ref{fig:mapping}, we compare FarmSeeker with the base segmentation model SegEarth-R1 and its sliding-window inference variants under different overlap ratios. All sliding-window variants use an input size of $512\times512$, with overlap ratios of 10\%, 30\%, and 50\%, respectively, to further evaluate the advantage of FarmSeeker over conventional local-context enhancement strategies. Subfigure~\ref{fig:mapping}(a) shows the mapping result of SegEarth-R1. Clear false positives and false negatives remain in several red-circled regions, particularly where farmland is adjacent to forests, roads, and built-up areas. In these complex regions, the model tends to produce missing boundaries, excessive mask expansion, and unstable local predictions. Subfigure~\ref{fig:mapping}(b) shows the result of FarmSeeker, which produces more complete and consistent farmland predictions in these regions and substantially alleviates the local errors of the base model. This indicates that FarmSeeker does not merely smooth the initial mask, but corrects erroneous decisions caused by insufficient observations through ambiguity-region identification, targeted querying, and collaborative reasoning. Subfigures~\ref{fig:mapping}(c)--(e) present the sliding-window inference results with overlap ratios of 10\%, 30\%, and 50\%, respectively. Although increasing the overlap ratio improves local segmentation continuity, the overall gains remain limited. Sliding-window inference mainly mitigates stitching discontinuities through repeated coverage, without changing the evidence used by the model to distinguish ambiguous regions. Consequently, clear false positives and false negatives remain in several red-circled regions even at higher overlap ratios. These results indicate that simply enlarging the local context or increasing window overlap cannot effectively resolve the deeper semantic ambiguities caused by phenological confusion, interference from neighboring land-cover types, and insufficient local observations.

\subsection{4.3. Comparison Between FarmSeeker and Non-expert Manual Farmland Interpretation in Real-world Ambiguous Scenarios}

To further evaluate the practical value of FarmSeeker in real-world complex scenarios, we compare it with non-expert manual farmland interpretation. Specifically, we select 10 real remote sensing image samples from the China subset of GSFS-Bench and recruit four undergraduate students without professional backgrounds in remote sensing as non-expert annotators. Expert farmland annotations are used as the ground truth to ensure the objectivity and consistency of the evaluation. In addition to segmentation accuracy, we record the annotation time of the non-experts and the inference time of FarmSeeker, thereby comparing their performance in real-world ambiguous scenarios from the perspectives of both accuracy and efficiency.

As shown in Table~\ref{tab:human_comparison}, FarmSeeker outperforms non-expert manual interpretation across all evaluation metrics. In particular, it identifies farmland regions more completely and effectively reduces the risk of missed detections in complex ambiguous areas. In scenarios where farmland boundaries are unclear and easily confused with roads, bare land, built-up areas, or post-harvest surfaces, non-experts tend to rely on intuitive judgment and are therefore more likely to omit local farmland regions. By contrast, FarmSeeker supplements the insufficient discriminative information in the current image through ambiguity perception, auxiliary evidence retrieval, and collaborative reasoning, thereby reducing omission errors. FarmSeeker also demonstrates a clear advantage in efficiency. The four non-experts require 734--1112~s to complete the annotation task, with an average time of approximately 897~s, whereas FarmSeeker completes the same task in only 243~s. These results indicate that FarmSeeker not only achieves higher segmentation accuracy but also substantially reduces the time cost. This advantage is particularly important for large-scale farmland mapping, where manual interpretation is time-consuming and often affected by differences in annotator experience.

\begin{table}[!t]
\centering
\resizebox{\columnwidth}{!}{
\begin{tabular}{lcccccc}
\toprule
Metric
& NE1
& NE2
& NE3
& NE4
& Mean
& FarmSeeker \\
\midrule
Recall $\uparrow$
& 81.21
& 80.52
& \underline{83.77}
& 74.94
& 80.11
& \textbf{89.68} \\

IoU $\uparrow$
& 72.02
& \underline{77.86}
& 76.70
& 64.10
& 72.67
& \textbf{78.26} \\

Acc. $\uparrow$
& 78.95
& \underline{83.08}
& 82.12
& 77.64
& 80.45
& \textbf{89.10} \\

Time (s) $\downarrow$
& 735
& \underline{734}
& 1112
& 1007
& 897
& \textbf{243} \\
\bottomrule
\end{tabular}
}
\caption{Comparison between FarmSeeker and non-expert manual farmland
interpretation in real-world ambiguous scenarios. NE denotes non-expert.}
\label{tab:human_comparison}
\end{table}

\begin{figure*}[!t]
\centering
\includegraphics[width=0.9\textwidth]{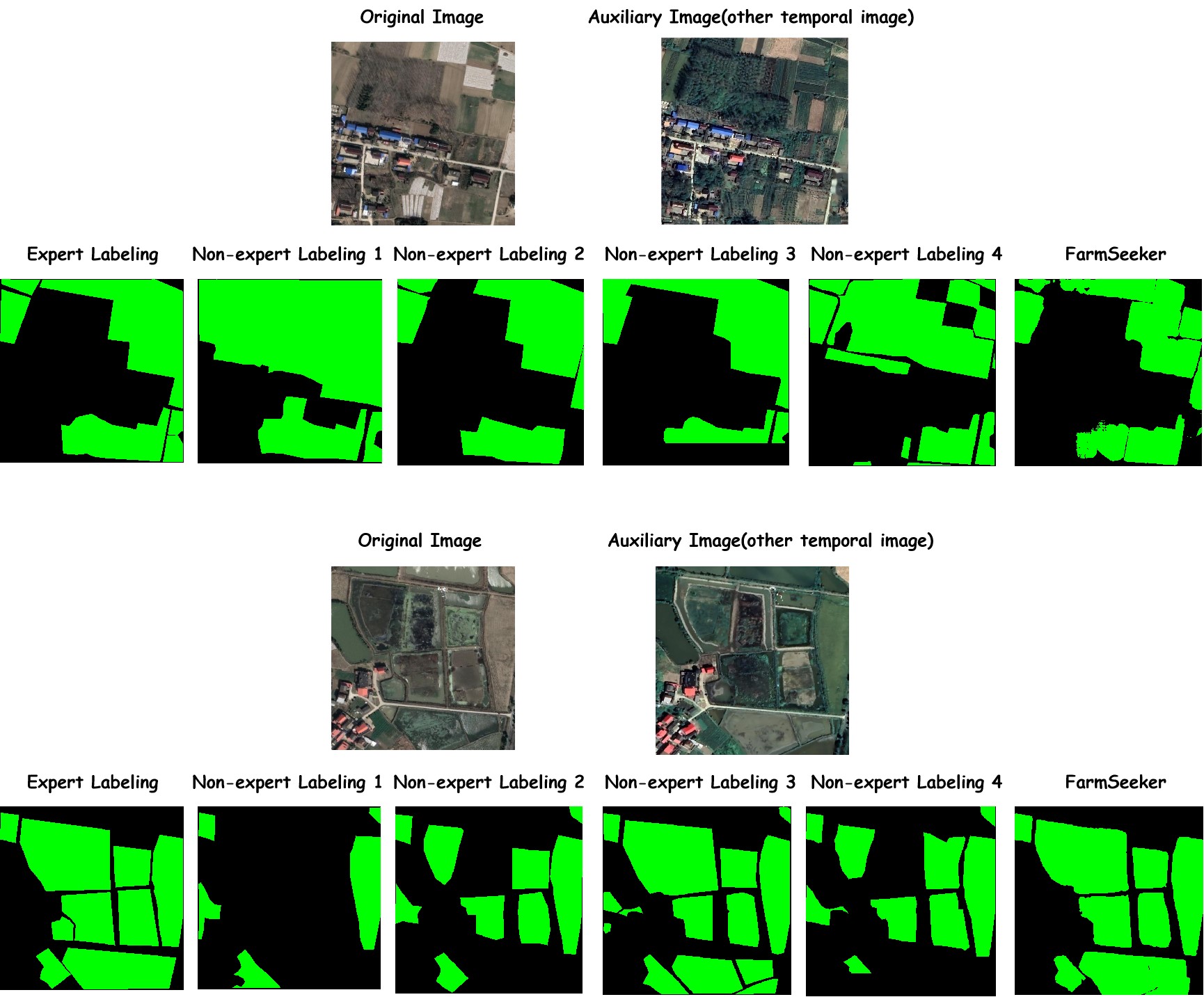} 
\caption{Comparison between FarmSeeker and non-expert manual farmland interpretation in representative real-world ambiguous scenarios. For each case, the original image, the auxiliary image from another timestamp, the expert labeling, the results from four non-expert annotators, and the prediction of FarmSeeker are shown.}
\label{fig:expert}
\end{figure*}

The qualitative results in Figure~\ref{fig:expert} further support these findings. In two representative ambiguous cases, the non-expert annotations show considerable inconsistencies along farmland boundaries and in complex background regions. Some annotations underestimate the farmland extent, whereas others fail to preserve the complete shapes of farmland parcels. In comparison, the segmentation results produced by FarmSeeker are generally closer to the expert annotations. With the assistance of auxiliary temporal imagery, FarmSeeker provides more reasonable and complete interpretations for regions that are difficult to identify from the original image alone. This indicates that FarmSeeker does not simply imitate manual annotation behavior. Instead, it surpasses non-expert manual interpretation in complex scenarios through ambiguity-driven dynamic evidence supplementation and collaborative reasoning.

Overall, the experiment demonstrates that FarmSeeker achieves higher segmentation accuracy, greater efficiency, and better consistency than non-expert manual interpretation in real-world ambiguous scenarios. These results further validate the practical potential of the proposed dynamic segmentation agent framework. FarmSeeker can serve as an efficient and stable tool for complex farmland interpretation, reducing the dependence of large-scale mapping tasks on manual interpretation experience.

\section{5. Step-by-step Visualization of FarmSeeker}
To provide a more intuitive analysis of the dynamic interpretation
process of FarmSeeker, this section randomly selects several cases and visualizes the complete execution chain from initial segmentation to final refinement. These cases mainly cover two typical scenarios, namely temporal ambiguity and insufficient spatial context.

\begin{figure}[!t]
\centering
\includegraphics[width=0.9\columnwidth]{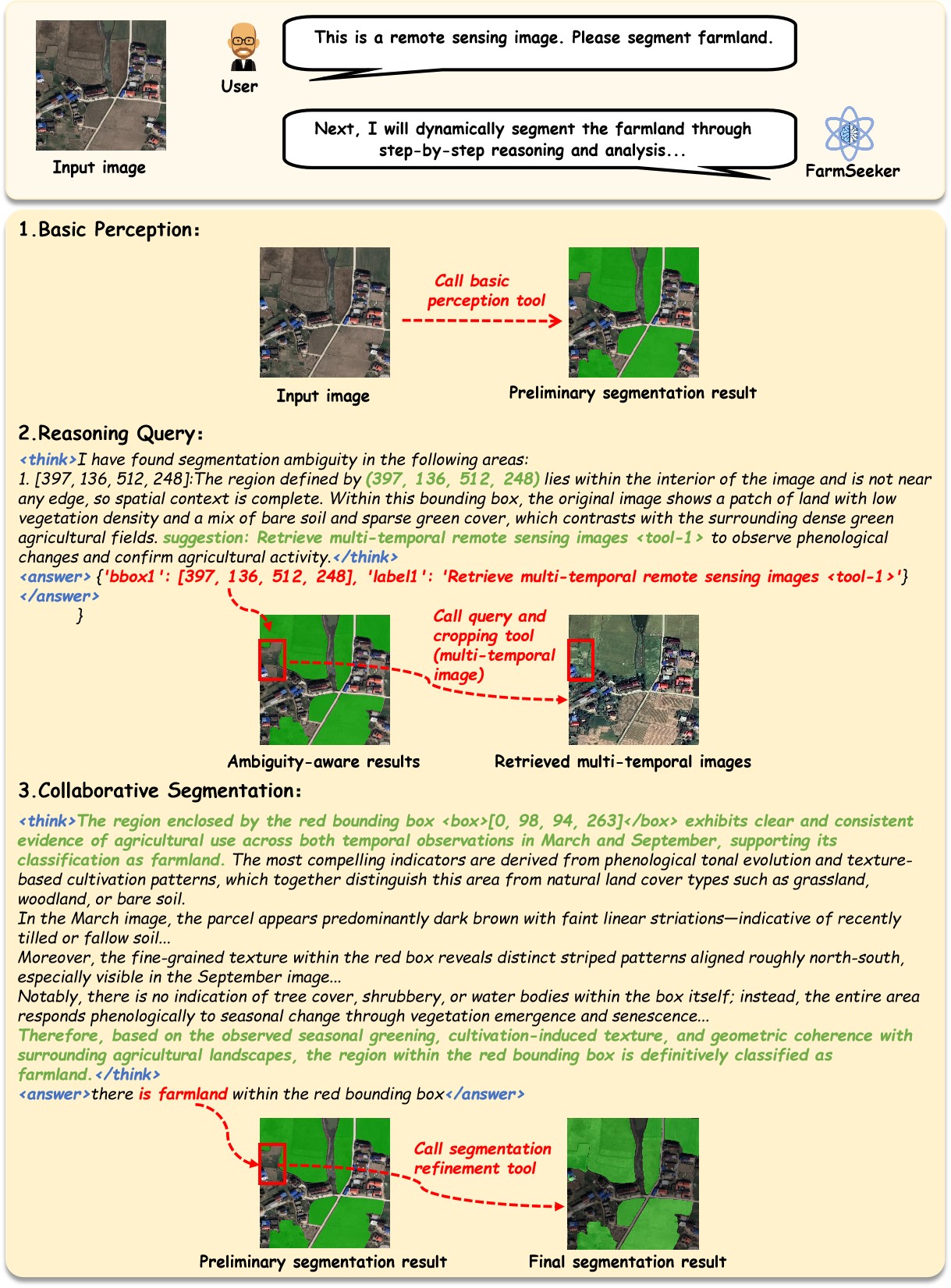} 
\caption{Example 1 of dynamic reasoning with multi-temporal auxiliary evidence.}
\label{fig:Case1}
\end{figure}

\begin{figure}[!t]
\centering
\includegraphics[width=0.9\columnwidth]{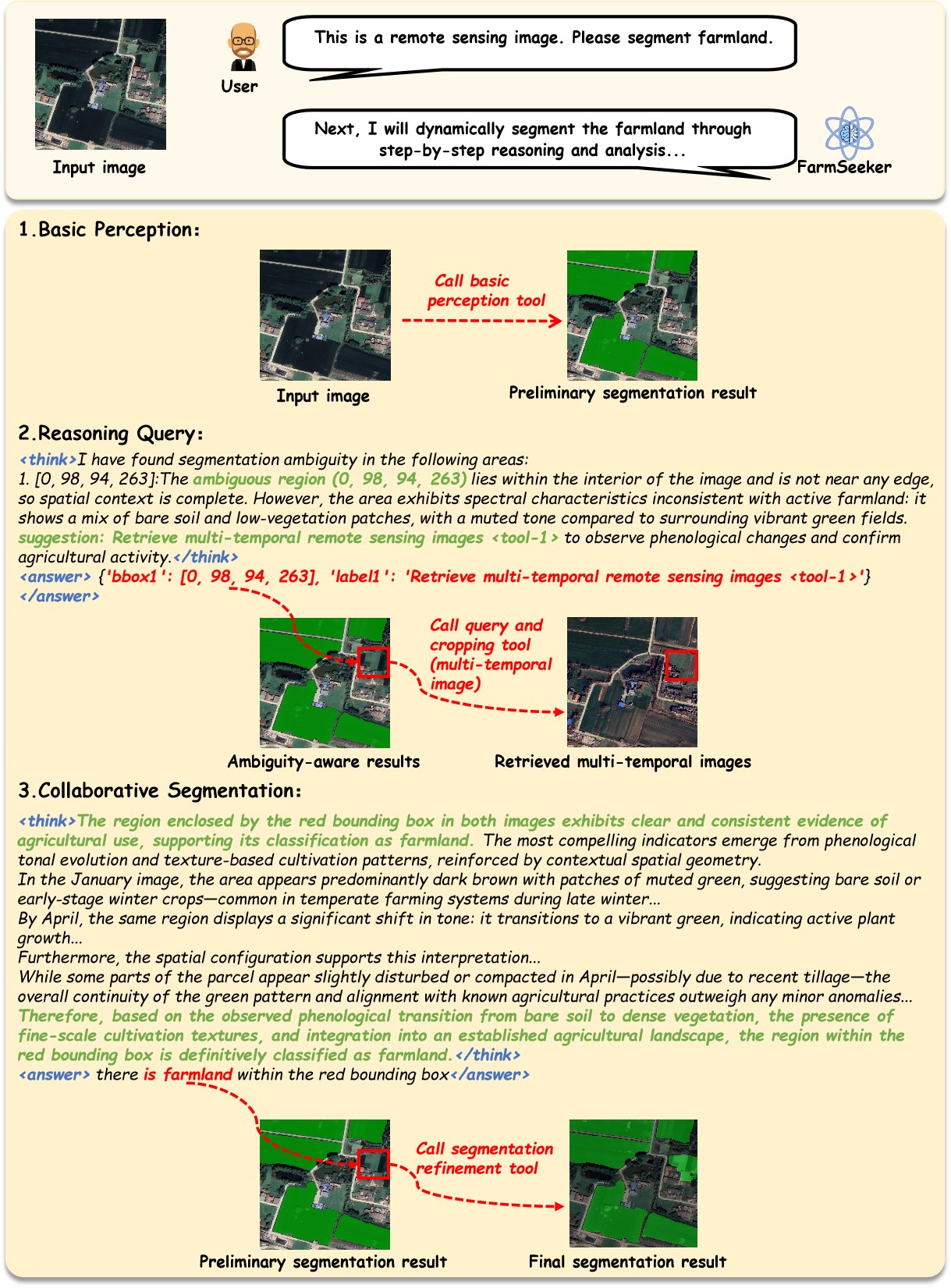} 
\caption{Example 2 of dynamic reasoning with multi-temporal auxiliary evidence.}
\label{fig:Case2}
\end{figure}

Figures~\ref{fig:Case1}-\ref{fig:Case2} illustrate the dynamic querying process driven by temporal ambiguity. In such scenarios, the target regions may be observed after harvest, during fallow periods, or at early crop growth stages, making their appearance more similar to bare soil or sparsely vegetated areas and consequently causing omissions in the initial segmentation. After identifying local anomalies in the initial mask, the reasoning engine determines that the current spatial context is sufficiently complete and therefore prioritizes the multi-temporal query tool. By comparing tonal changes, texture evolution, and temporal consistency with the surrounding farmland across different months, the system confirms that the target regions exhibit stable agricultural activity patterns and corrects them as farmland. This process demonstrates that when ambiguity mainly arises from phenological changes and temporal appearance variations, additional temporal evidence can effectively compensate for insufficient single-temporal observations.

\begin{figure}[!t]
\centering
\includegraphics[width=0.9\columnwidth]{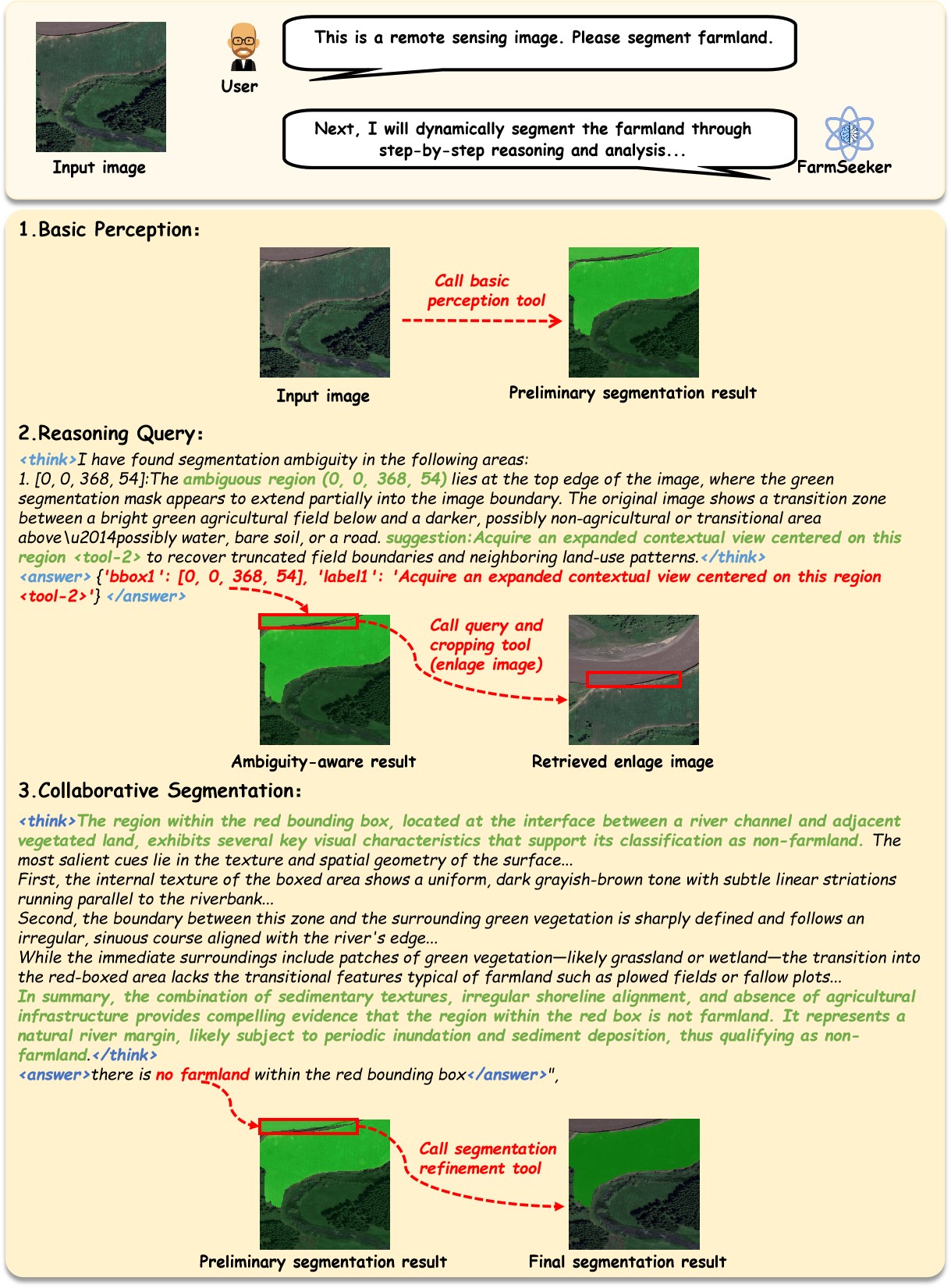} 
\caption{Example 1 of dynamic reasoning with enlarged image retrieval.}
\label{fig:Case3}
\end{figure}

\begin{figure}[!t]
\centering
\includegraphics[width=0.9\columnwidth]{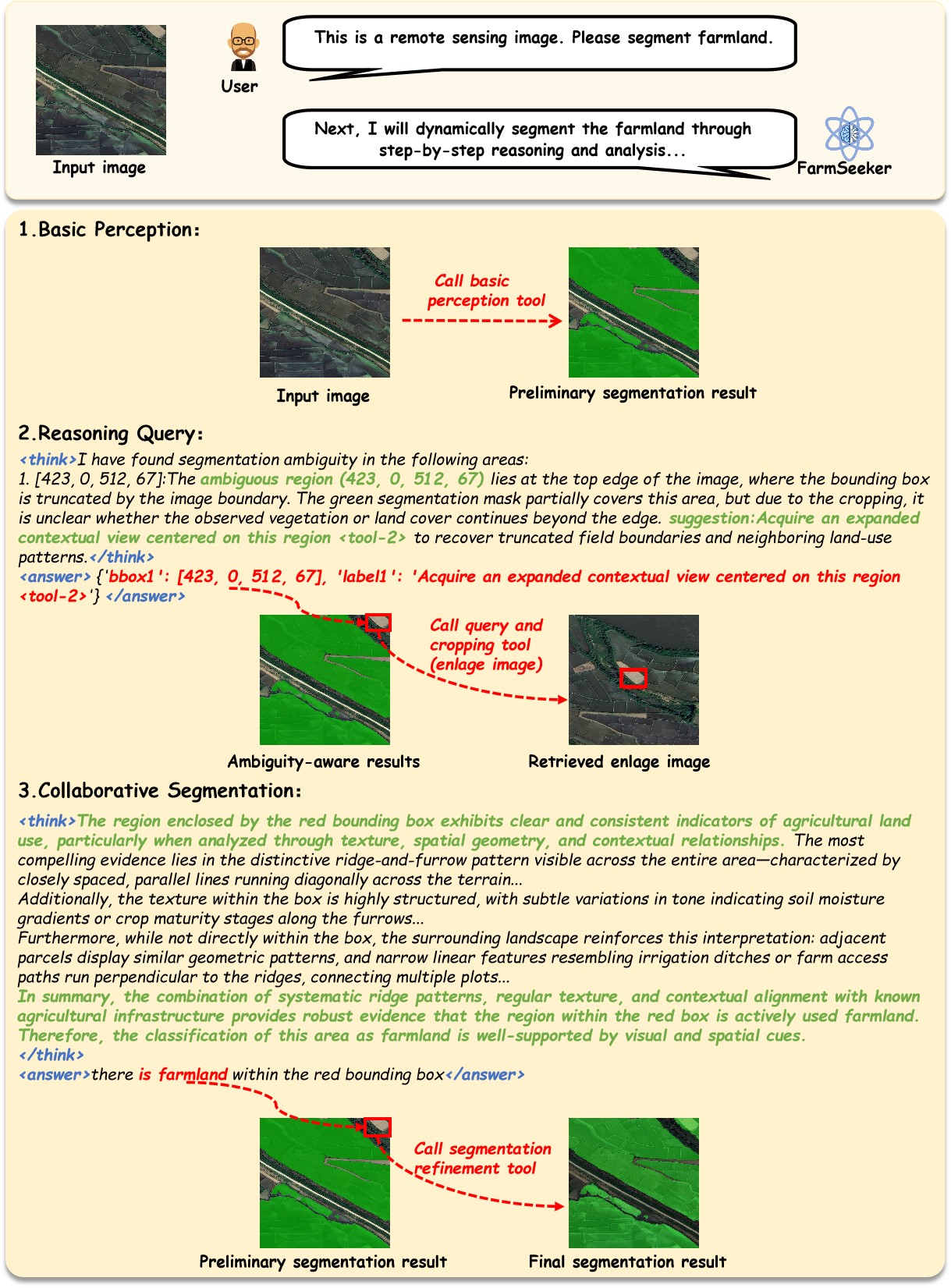} 
\caption{Example 2 of dynamic reasoning with enlarged image retrieval.}
\label{fig:Case4}
\end{figure}

Figures~\ref{fig:Case3}-\ref{fig:Case4} illustrate the dynamic querying process driven by insufficient spatial context. Such ambiguous regions are typically located near image edges or cropping boundaries. Because the target structures are truncated, the model cannot reliably determine their true properties using only the current field of view. After detecting this type of boundary ambiguity, the reasoning engine selects the enlarge query tool to retrieve a broader spatial context from the wide-swath image and further analyzes the structural continuity, texture arrangement, and geometric relationships between the target region and surrounding objects. In Figure~\ref{fig:Case3}, the enlarged view shows that the target region lies at the boundary between a river channel and surrounding natural surfaces and lacks typical farmland textures; it is therefore corrected as non-farmland. In Figure~\ref{fig:Case4}, the target region exhibits clear continuity with surrounding regular fields in ridge texture, orientation, and spatial organization and is therefore restored as farmland. This process shows that expanding the spatial context can effectively recover missing evidence caused by a limited local field of view or boundary truncation.

These cases further demonstrate that the advantage of FarmSeeker does not come from simply introducing more images, but from selecting auxiliary evidence with greater information gain according to the source of ambiguity.

\end{document}